\documentclass[manuscript,screen,nonacm]{acmart}

\AtBeginDocument{%
  }

\setcopyright{acmlicensed}
\copyrightyear{2025}
\acmYear{2025}
\acmDOI{XXXXXXX.XXXXXXX}
\acmConference[Conference acronym 'XX]{Make sure to enter the correct
  conference title from your rights confirmation email}{June 03--05,
  2018}{Woodstock, NY}
\acmISBN{978-1-4503-XXXX-X/2018/06}

\usepackage{graphicx}%
\usepackage{multirow}%
\usepackage{subcaption} 
\usepackage{amsthm}%
\usepackage{mathrsfs}%
\usepackage[title]{appendix}%
\usepackage{xcolor}%
\usepackage{textcomp}%
\usepackage{manyfoot}%
\usepackage{booktabs}%
\usepackage{algorithm}%
\usepackage{algorithmicx}%
\usepackage{algpseudocode}%
\usepackage{listings}%
\usepackage[export]{adjustbox} 
\usepackage{longtable}
\usepackage{booktabs}
\usepackage{array}
\usepackage{ragged2e}
\usepackage{setspace}
\usepackage[most]{tcolorbox}
\tcbset{colback=gray!10, colframe=gray!50, boxrule=0.5pt, arc=2mm}

\begin{document}

\title[What Questions Should Robots Be Able to Answer?]{What Questions Should Robots Be Able to Answer? A Dataset of User Questions for Explainable Robotics}

\author{Lennart Wachowiak}
\email{lennart.wachowiak@kcl.ac.uk}
\orcid{0000-0001-6105-609X}
\affiliation{%
\city{London}
  \institution{King's College London, CDT in Safe and Trusted AI}
  \country{UK}
}

\author{Andrew Coles}
    \orcid{0000-0002-4954-9235}

\affiliation{%
  \institution{King's College London}
  \city{London}
  \country{UK}
  }

\author{Gerard Canal}
  \orcid{0000-0002-6718-1198}

\affiliation{%
  \institution{King's College London}
  \city{London}
  \country{UK}}

\author{Oya Celiktutan}
\orcid{0000-0002-7213-6359}

\affiliation{%
 \institution{King's College London}
 \city{London}
 \country{UK}
 }

\renewcommand{\shortauthors}{Wachowiak et al.}

\begin{abstract}
With the growing use of large language models and conversational interfaces in human--robot interaction, robots' ability to answer user questions is more important than ever.
We therefore introduce a dataset of 1,893 user questions for household robots, collected from 100 participants and organized into 12 categories and 70 subcategories. Most work in explainable robotics focuses on 
why-questions, such as ``Why did you clean the bathroom but not the kitchen?'' In contrast, our dataset provides a wide variety of questions, from questions about simple execution details to questions about how the robot would act in hypothetical scenarios --- thus giving roboticists valuable insights into what questions their robot needs to be able to answer.
To collect the dataset, we created 15 video stimuli and 7 text stimuli, depicting robots performing varied household tasks. We then asked participants on Prolific what questions they would want to ask the robot in each portrayed situation.  
In the final dataset, the most frequent categories are questions about task execution details (21.4\%), the robot's capabilities (12.6\%), and performance assessments (10.7\%). 
Although questions about how robots would handle potentially difficult scenarios and ensure correct behavior are less frequent, users rank them as the most important for robots to be able to answer.
Moreover, we find that users who identify as novices in robotics ask different questions than more experienced users. Novices are more likely to inquire about simple facts, such as what the robot did or the current state of the environment. 
As robots enter environments shared with humans and language becomes central to giving instructions and interaction, this dataset provides a valuable foundation for (i) identifying the information robots need to log and expose to conversational interfaces, (ii) benchmarking question-answering modules, and (iii) designing explanation strategies that align with user expectations.
\end{abstract}




\maketitle

\section{Introduction}
As robots enter homes, healthcare settings, and service environments \cite{trainum2023robots, gonzalez2021service}, where they perform tasks for or with people, natural language communication becomes essential. Language now also plays a central role in instructing robots; especially with vision--language--action models (VLAs), which bring large language models (LLMs) to the physical domain and map visual observations and language instructions to robot actions~\cite{team2025gemini}. Widely publicized VLA demos often feature household tasks, such as cleaning or preparing meals --- areas with ample human contact. 

When interacting with such a robot, users might naturally have many questions. Users may want to understand how the robot works, what its decision-making process looks like, and how it would deal with unexpected issues. A robot's ability to explain itself in such situations fosters safer interactions and is known to help users calibrate their trust levels appropriately~\cite{wangtrust2016}. With advances in text generation capabilities \cite{achiam2023gpt, comanici2025gemini}, answering user questions is easier than ever. However, for the answers to be truthful, they must be based on the robot's true capabilities, underlying decision-making, and execution traces. If the robot's question-answering module lacks this data, an LLM might simply confabulate the answers \cite{huang2025survey}. Without relevant logs and access to appropriate explainability methods, the LLM's answers may devolve into statistically likely language patterns unrelated to the robot's actual behavior. Therefore, roboticists need to know what data should be logged and made accessible to such a question-answering module. Yet, despite growing interest in explainable robotics, we still lack a broad empirical account of what users actually want to ask robots across everyday tasks. In this paper, we present a novel large-scale dataset of questions users have for household robots. 

Such a dataset has several uses for explainable robotics and human--robot interaction. First, it helps roboticists anticipate what kinds of questions users may expect a household robot to answer. Second, it clarifies what information robots may need to log during task execution and expose to a natural language interface, such as action histories, object states, capability boundaries, or sources of knowledge. 
Third, the dataset and category hierarchy can support the design and benchmarking of XAI and question-answering systems, which we demonstrate in our follow-up work \cite{wachowiak2026neurosymbolic}.

Our analysis focuses on general-purpose robots executing household tasks as these offer desirable features for our research: close human contact, a need for natural language communication, and tasks people would like robots to perform for them \cite{li2023behavior}. Recent research and industry demonstrations are shifting from robots with narrow capabilities to robots that can execute a wide range of household tasks. 
This shift is reflected in our question categorization, which revolves around questions that recur across tasks and settings. This does not mean that task context is irrelevant: each question in the dataset remains linked to the stimulus from which it was collected, enabling future task-conditioned analyses.

Beyond providing the dataset, our paper addresses the following research questions in an exploratory manner:
\begin{itemize}
    \item \textbf{RQ1}: What types of user questions should service robots be able to answer?
    \item \textbf{RQ2}: What types of questions do users consider especially important?
    \item \textbf{RQ3}: Do the types of questions and the importance the user places on the robot's question-answering capabilities vary with the 
    user's robot experience and attitudes towards robots?
\end{itemize}

Our main findings and contributions can be summarized as follows:
\begin{itemize}
    \item \textbf{We provide a dataset of 1,893 user questions. We have hierarchically categorized the questions into 12 main categories and 70 subcategories (RQ1)}. The resulting categories show a wide variety of question types, many of which are underexplored in the current explainable robotics literature. Beyond the causal why-questions often explored in XAI, the categories include questions about execution details, the robot's capabilities, whether and how it can handle potential difficulties, technical details of the robot, its sources of knowledge, and many more. 
    \item \textbf{We find that users assign different levels of importance to different question categories (RQ2)}. However, a unifying theme is that questions about safety, failure, and potential issues rank highly. Questions about how the robot would deal with hypothetical scenarios and how it ensures its actions are correct rank highest. In contrast, the why-questions popular in XAI literature have the second-lowest average importance score. 
    \item \textbf{We find differences by user characteristics (RQ3).} The types of questions asked differ based on a person's level of experience with robots. Novices allocate a larger share of their questions to ask about execution details and the state of the environment. In contrast, more experienced users are more likely to ask questions about how the robot would deal with potentially difficult situations or how it makes decisions. Moreover, users with more positive attitudes towards robots also think it is more important for robots to be able to answer their questions. These findings highlight the importance of adaptive, user-centered explainability that takes users' prior experience and expectations into account. 
\end{itemize}

\section{Related Work}

Our work contributes to the field of explainable robotics. Therefore, Section \ref{sec:rel-xai} introduces the field, with a special focus on explainability research that focuses on answering natural language questions, often with the help of LLMs. Section \ref{sec:rel-user-dif} highlights how explanation needs can vary between users. Section~\ref{sec:rel-hri-data} introduces other datasets relevant for conversational HRI. Lastly, Section \ref{sec:rel-question-cat} introduces other work that categorizes user questions for AI and robots. 

\subsection{Explainable AI/Robotics} \label{sec:rel-xai}
Explainability has become a core topic in the robotics \cite{anjomshoae2019explainable, setchi2020explainable}, AI planning \cite{fox2017explainable}, and machine learning \cite{8466590} literature. By providing the reasons for the robot's actions and decisions, explanations are supposed to:
\begin{itemize}
    \item enable users to understand the robot's decision-making and be better at predicting future behavior \cite{wachowiak2023survey, doshi2017towards},
    \item allow users to calibrate their trust levels appropriately \cite{fischer2018increasing, lyons2023explanations, wang2016impact, anjomshoae2019explainable},
    \item  improve human--agent team performance due to enhanced knowledge transfer \cite{wang2016impact},
    \item  and improve how an agent is perceived \cite{choi2021err}. 
\end{itemize}
A myriad of XAI modules for robots and agents have been developed in the past, typically focusing on specific functionalities of the robot, such as explainable vision~\cite{buhrmester2021analysis}, classical planning \cite{fox2017explainable, cashmore2019towards}, or motion modules \cite{9562003}. As in our work, previous research has also emphasized the importance of considering user needs and requirements when developing XAI modules, for example, through participatory design \cite{mucha2020towards, eiband2018bringing, ehsan10.1007/978-3-030-60117-1_33, wachowiak2023survey}. For instance, Gebellí et al. ran a co-design study in a geriatric unit to understand how to meet the needs of nurses in making a robot legible and explainable \cite{gebelli2024co}.

With the rise of LLMs (e.g., \cite{achiam2023gpt, guo2025deepseek}), these explainability modules can interface with users through natural language. 
LLMs are now used to answer various user questions by retrieving relevant information from logs or specific XAI modules and synthesizing answers. For example, Sobrín-Hidalgo et al. \cite{sobrin2024explaining} propose to collect the execution logs of a robot module in a vector database. Once the robot is asked to answer an explanation-seeking question, an LLM provides the answer using Retrieval Augmented Generation (RAG), thus querying the relevant bits from the database and summarizing the logs for the user. Moreover, Gebellí et al. \cite{gebelli2025personalised} use LLMs to build a user profile based on past interactions so that the generated explanations align more closely with that user's mental model. Lastly, LeMasurier et al. \cite{lemasurier2024templated} compare LLM-generated with templated explanations, for a robot controlled via a behavior tree \cite{tagliamonte2024generalizable}.

However, the user questions that can be answered depend on the information available to the LLM. Being aware of all possible user questions, their importance, and frequency is crucial for developing robots that log the right data and for implementing relevant XAI methods. Allowing LLMs to ground their answers in actual facts is especially important considering the well-known LLM problem of hallucinations/confabulations \cite{huang2025survey}. Our dataset addresses this challenge by helping roboticists better understand users' information needs and guiding practitioners in what robot systems should record and expose to downstream question-answering modules. 

\subsection{User Differences in Explanation Needs} \label{sec:rel-user-dif}
Different types of users (e.g., based on cognitive styles or socio-cultural backgrounds \cite{kopecka2024preferences, kopecka2020explainable}) might have varying explanation needs and preferences; changing task circumstances can further complicate explanation needs \cite{wachowiak2024people, wachowiak2024large}. For instance, explanations have been found to be more effective at repairing trust when a person's attitude towards robots is more positive \cite{esterwood2022having}. The researchers explain this phenomenon by stating that explanations can offer a rationale or justification for a user's positive attitudes, thus reducing the user's potential to experience cognitive dissonance after being confronted with a robot error. Given the findings in previous research, we decided to explore whether a person with more positive attitudes towards a robot, as measured by the GAToRS questionnaire \cite{koverola2022general}, also thinks it is more important for a robot to be able to answer their questions. Generally, people who see robots more positively might also expect more communicative capabilities from robots (to not fall into a state of dissonance with their positive pre-conceptions), as well as simply show a greater interest in their communicative capabilities. On the other hand, users with more negative attitudes might especially be interested in explanation capabilities as a tool to build trust. Moreover, we explore the influence of the understudied variable of the user's experience with robots.

\subsection{Datasets for HRI}  \label{sec:rel-hri-data}

There is a wide variety of datasets aimed at advancing various aspects of HRI. Such datasets encompass both natural language communication and multimodal non-verbal communication. Non-verbal datasets include recorded human--human interactions \cite{ji2019survey, ko2021air} and human--robot interactions \cite{spitale2024err, candon2024react, gucsi2025hri}, for example, focused on understanding users' social cues or facilitating the generation of human-like gestures.
Examples of datasets that deal with natural language communication focus on user instructions and their translation into robot actions \cite{shridhar2020alfred, nair2022learning, bastianelli2014huric}. Others focus on robots that ask clarification questions to better understand their task \cite{gao2022dialfred}; while containing questions, these are from the robot to the user, thus in the opposite direction compared to the questions we focus on in our research. 

Most relevant to our research, there is a small set of natural language corpora of HRI-relevant dialogues. For instance, the SCOUT corpus \cite{lukin-etal-2024-scout} consists of users instructing a remote robot to explore its environment. The Vernissage corpus \cite{jayagopi2013vernissage} consists of conversations in which a robot talks about paintings and subsequently quizzes people about them. Lastly, the TEACh corpus \cite{padmakumar2022teach} contains human--human dialogues on completing household tasks, with one person taking the role of the ``commander'' and another one the role of the ``follower''. However, these corpora and interactions are not designed to analyze users' questions for the robot --- the focus of our paper. Research and datasets on user questions are highlighted in the next section. 

\subsection{Question Categorizations} \label{sec:rel-question-cat}
In our previous work, we categorized the robot modules (e.g., navigation, manipulation, vision, etc.) for which explainability modules have been developed in the explainable robotics literature \cite{wachowiak2024taxonomy}. 
Furthermore, we provided a categorization of situations, often caused by an interaction rupture or unexpected event, in which users demand an explanation from the robot  (e.g., robot errors, uncertainty, suboptimal behavior) \cite{wachowiak2024people}. While these taxonomies provide us with insights into the HRI contexts in which questions are asked, in this work, we present a dataset of concrete natural language questions users ask in a variety of household-related robotics tasks.

McGuinness et al. \cite{mcguinness2007categorization} present a categorization of user questions in the context of a software agent supporting the purchase of a laptop. However, the resulting taxonomy is less detailed than what we present in the following; they provide no dataset of natural language questions, and the topic is not concerned with robotics.  
Raza et al. \cite{raza2022wild} explore what questions users ask robots in the context of users encountering a robot at a university's open day. The 179 questions asked by users were either standard conversation openers (e.g., ``How are you?''), or were concerned with facts and opinions about the university, general facts, or the robot itself (personality, beliefs, goals, abilities). In contrast, our study focuses on users encountering a robot executing a specific task, rather than a social small talk setting. We prompt the users to ask questions related to the robot in the task context, as we are not interested in creating a dataset of questions in the context of social chit-chat, such as ``What is your favorite movie?'' or ``How is the weather today?'' Lastly, our study provides a dataset that is more than 10 times larger and is coded into categories relevant to researchers building XAI and question-answering modules, enabling users to better understand the robot. 

\section{Method --- How the Dataset was Collected}

\subsection{Study Procedure}
Participants were recruited on Prolific, from where they accessed our questionnaire created in Qualtrics. The questionnaire form is publicly available (see Appendix \ref{app:online}).
After reading an information sheet and consent form, each participant was required to watch four or five video scenarios and read two text scenarios. The stimuli order was randomized. Videos and texts were randomly selected out of a larger pool of 22 stimuli described in Section \ref{sec:stimuli}.
After encountering a video or text scenario, users had to provide natural language questions they would want to ask the robot. Specifically, users answered the following prompt: 
\begin{tcolorbox}
What questions would you want the robot to be able to answer about its behavior related to the task?  
Please provide questions directly related to what you saw in the video.
Importantly, we are looking for questions in response to which the robot has to explain itself and its behavior. We do not look for general questions about the world, such as ``How is the weather?''
\end{tcolorbox}

Based on this prompt, each user had to provide two questions. We collected two further questions with the following prompt: 
\begin{tcolorbox}
    Beyond questions related directly to the video, what type of questions about its task and behavior do you think a robot like in the video should be able to answer?
\end{tcolorbox}

In addition, users also provided importance scores for their questions, answering the following question on a Likert scale from 1 (Not at all important) to 5 (extremely important):
\begin{tcolorbox}
How important is it to you that a robot you interact with can answer such a question?
\end{tcolorbox}

In the end, users provided some demographic data and filled out a 10-item questionnaire on people's general attitudes towards robots. The attitude scores are measured via the recent, validated GAToRS questionnaire \cite{koverola2022general}, using the two positive subscales (7-point Likert scales) for personal-level and societal-level attitudes towards robots. An example of a statement probing societal-level attitudes towards robots is ``Robots are a good thing for society, because they help people''. An example of a statement probing personal-level attitudes towards robots is ``I would feel relaxed talking with a robot''.

\subsection{Video and Text Stimuli} \label{sec:stimuli}
To determine the types of questions users have for robots assisting them with everyday tasks, we present users with video or text scenarios. Videos and texts were selected to cover a large diversity of tasks, robot embodiments, and interaction settings. The featured tasks include a variety of common household tasks, such as cleaning, washing dishes, doing laundry, and cooking. The featured robot embodiments include a variety of humanoids and mobile manipulators. Regarding the interaction setting, the videos feature robots by themselves, interacting with humans, and in multi-robot setups. The household tasks included in the text scenarios were inspired by the tasks people reported in previous research as those they would want robots to perform \cite{li2023behavior}.

\paragraph{Videos}
The videos depict robots accomplishing multistep tasks in a home, for example, making the bed and then removing trash from the floor. Table \ref{tab:video_overview} describes each video's content and provides an example frame. The videos were sourced from publicly available robot demos, either teleoperated or featuring state-of-the-art policies. The original videos are cut to show the execution of one self-contained task. Specifically, we included videos from Google's Gemini Robotics \cite{team2025gemini}, Physical Intelligence \cite{intelligence2025pi_}, the Mobile ALOHA project \cite{fu2024mobile}, Figure\footnote{\url{https://www.figure.ai/}}, Astribot\footnote{\url{https://www.astribot.com/en}}, Hello Robot\footnote{\url{https://hello-robot.com/}}, and 1X\footnote{\url{https://www.1x.tech/}}. 
The average video length was 35.4 seconds (SD=13.8). This short length is appropriate for an online user study, while still allowing us to show longer tasks, such as preparing a meal, by cutting between key moments of the task execution.
Links to the original, uncut videos can be found in Appendix \ref{app:video-ref}.

\renewcommand{\arraystretch}{1}  
\begin{table*}[!htbp]
     \caption{Overview of the video stimuli used. The links to the original videos can be found in Appendix \ref{app:video-ref}.}
         \label{tab:video_overview}

    \centering
    \begin{adjustbox}{width=\textwidth}
    \begin{tabular}{c|p{8cm}|p{4cm}|c}  
    \textbf{Example Frame}&\textbf{Video Description}&\textbf{Robot}&\textbf{Length} \\
    \hline
        \raisebox{-0.5\height}{\includegraphics[width=2.5cm, cfbox=black 0.6pt 0pt]{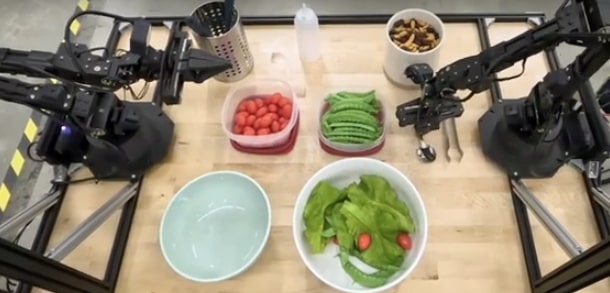}}  &  The robot finishes a salad bowl by adding nuts and peas using a spoon and pliers [V1]. &   ALOHA-2 (Gemini-Robotics)& 35s\\

       \raisebox{-0.5\height}{\includegraphics[width=2.5cm, cfbox=black 0.6pt 0pt]{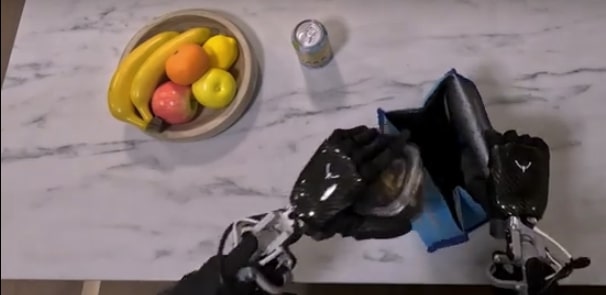}}  & The robot packs a lunchbox with a nut mix and an orange [V2].   &  Apollo (Gemini-Robotics)  & 19s \\

      \raisebox{-0.5\height}{\includegraphics[width=2.5cm, cfbox=black 0.6pt 0pt]{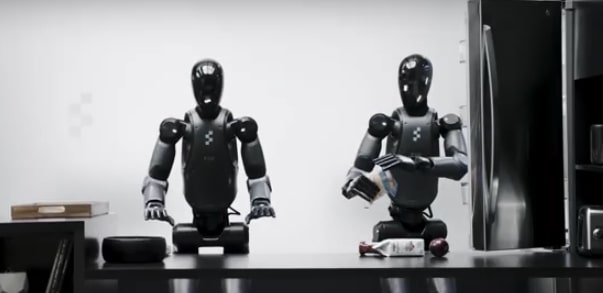}}  & Two robots collaborate on putting groceries away into the fridge and a wooden box [V3].   &  Figure 02  & 43s \\

      \raisebox{-0.5\height}{\includegraphics[width=2.5cm, cfbox=black 0.6pt 0pt]{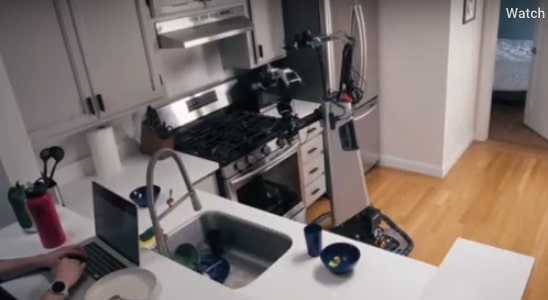}}  & The robot closes a cupboard, cleans a spill, and puts away dishes into the sink [V4].   &  Mobile Manipulator from Physical Intelligence  & 67s \\

    \raisebox{-0.5\height}{\includegraphics[width=2.5cm, cfbox=black 0.6pt 0pt]{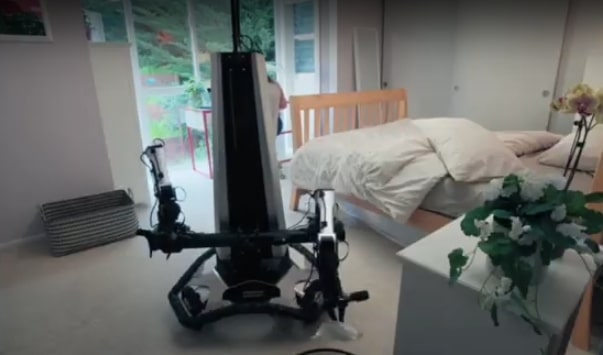}}  & The robot tidies up a bedroom by putting away clothes, removing trash, and making the bed [V5].   &  Mobile Manipulator from Physical Intelligence  & 25s \\

    \raisebox{-0.5\height}{\includegraphics[width=2.5cm, cfbox=black 0.6pt 0pt]{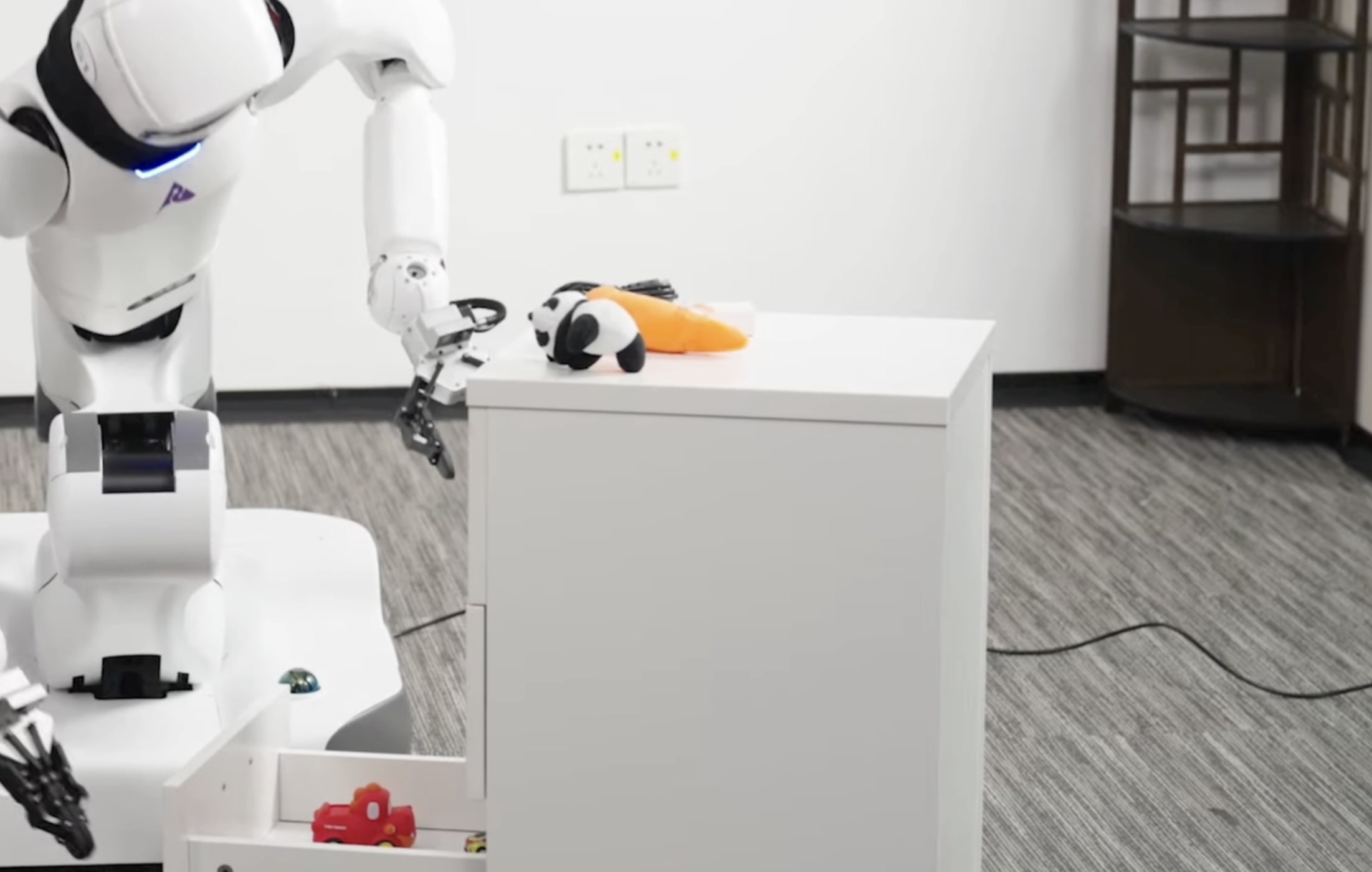}}  & The robot sorts loose items into different shelves [V6].   &  Astribot S1  & 33s \\

    \raisebox{-0.5\height}{\includegraphics[width=2.5cm, cfbox=black 0.6pt 0pt]{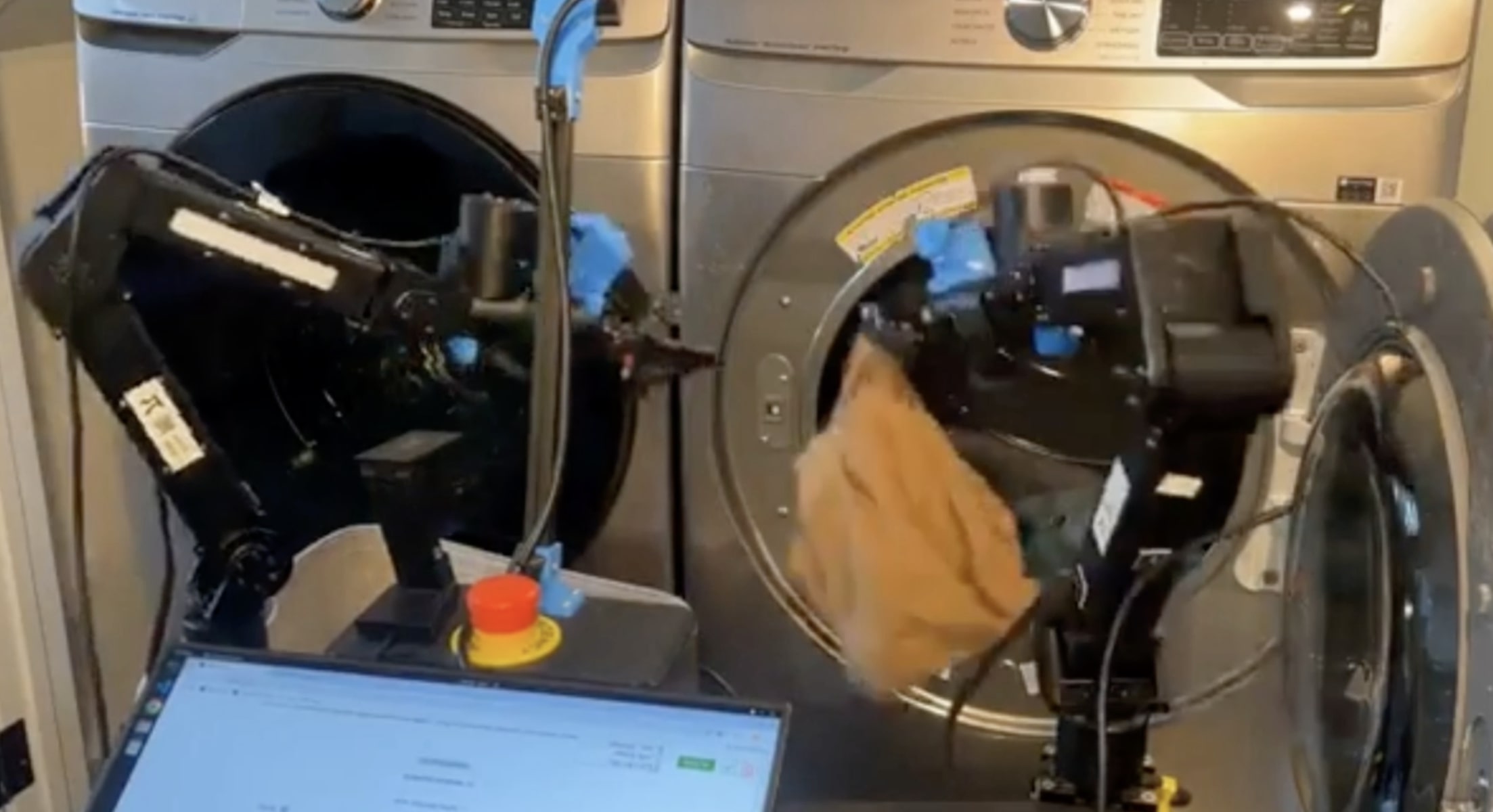}}  & The robot unpacks clothes from a dryer into a basket [V7].   &  Mobile Manipulator from Physical Intelligence  & 41s \\

    \raisebox{-0.5\height}{\includegraphics[width=2.5cm, cfbox=black 0.6pt 0pt]{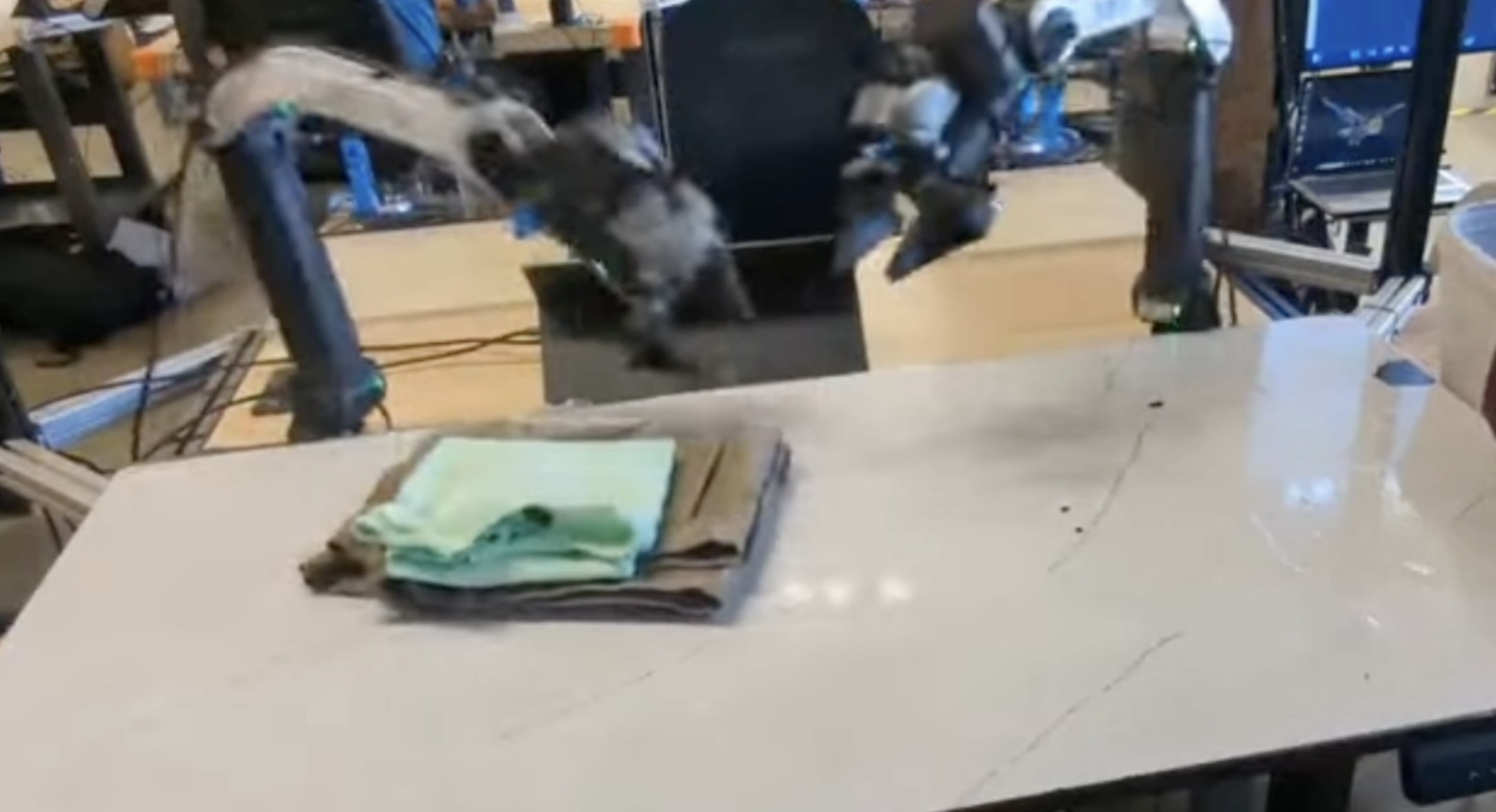}}  & The robot folds and stacks three items of clothes [V8].   &  Mobile Manipulator from Physical Intelligence  & 59s \\

    \raisebox{-0.5\height}{\includegraphics[width=2.5cm, cfbox=black 0.6pt 0pt]{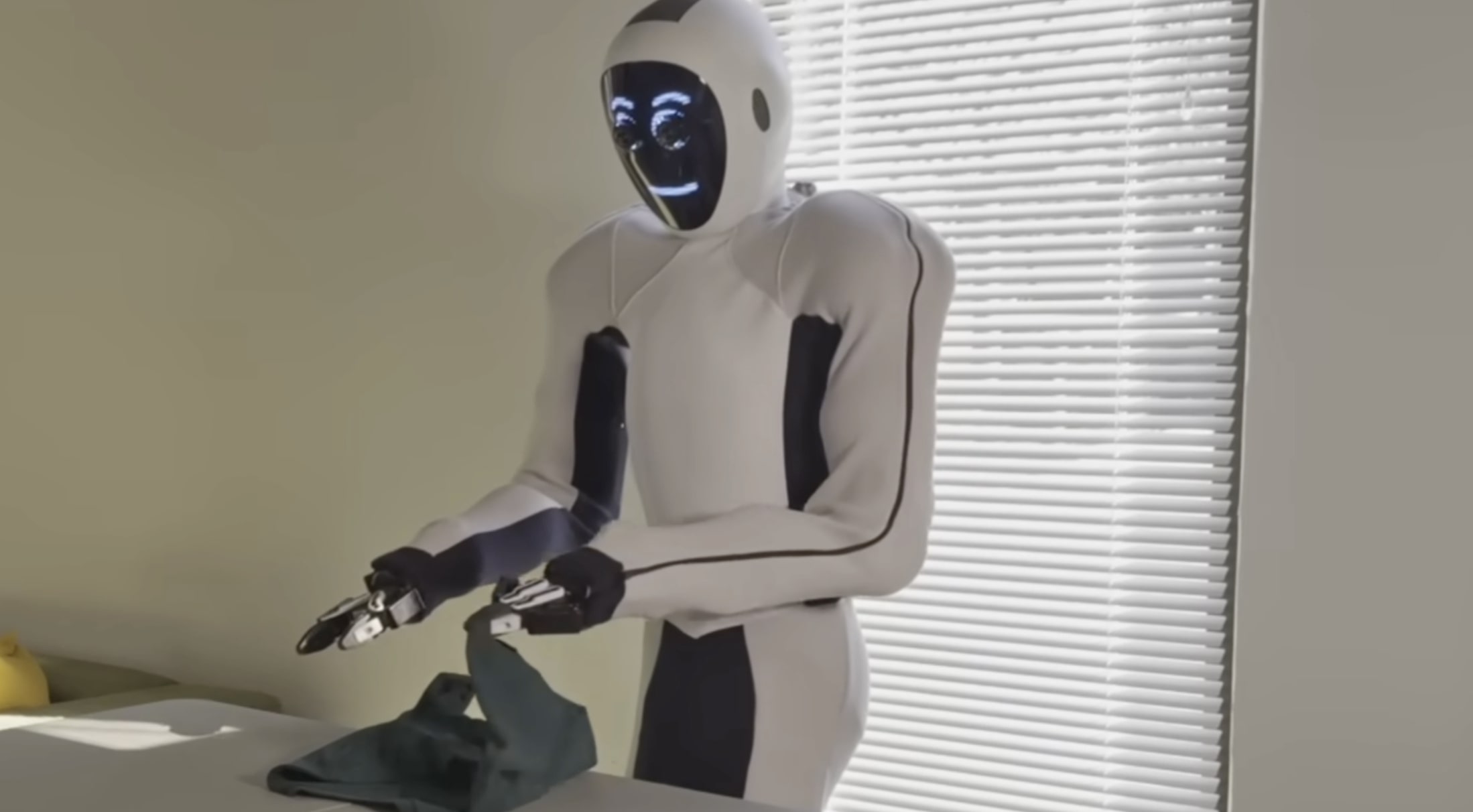}}  & The robot folds a t-shirt and hands it over to a person [V9].   &  1X Eve  & 27s \\

    \raisebox{-0.5\height}{\includegraphics[width=2.5cm, cfbox=black 0.6pt 0pt]{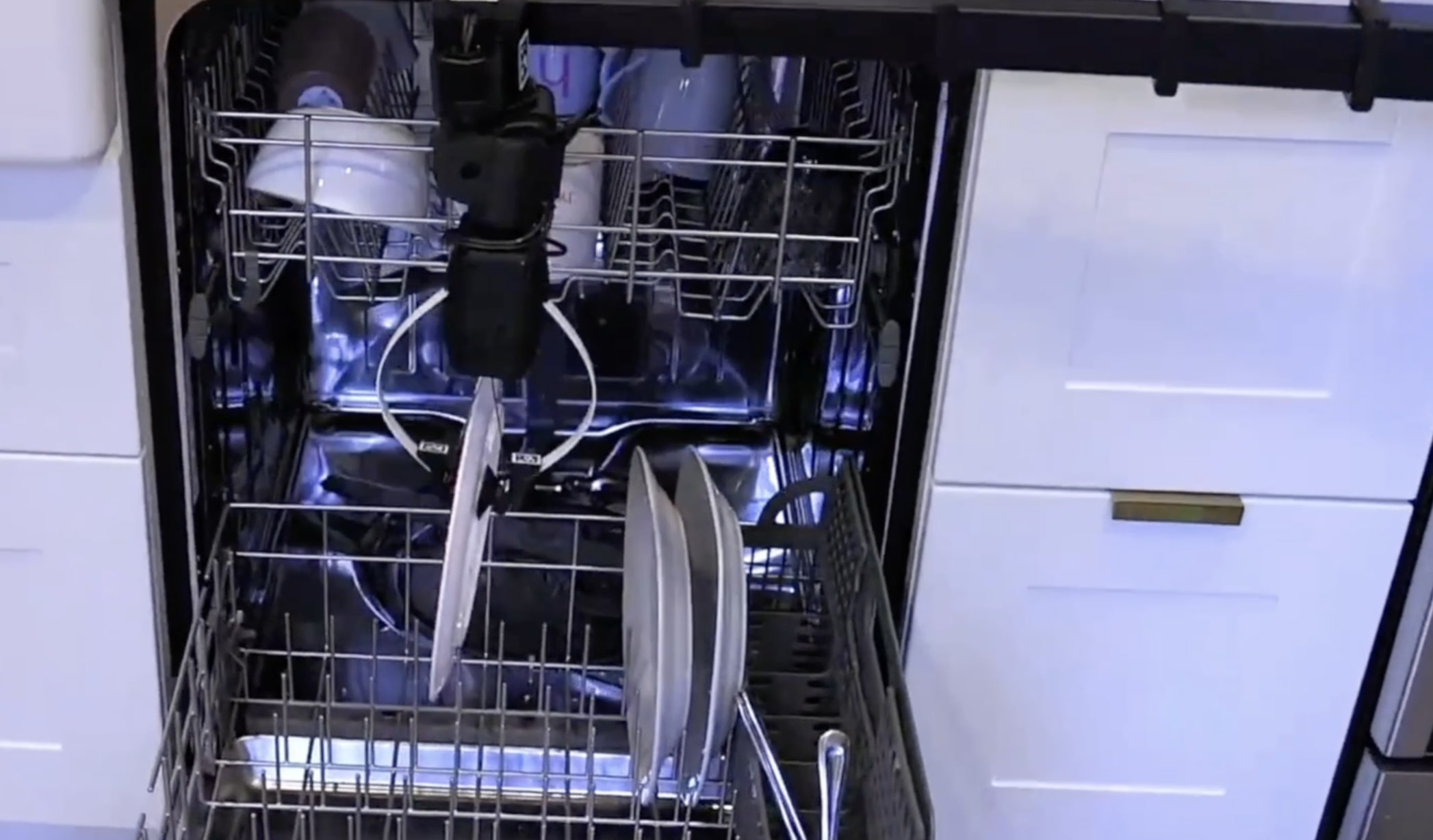}}  & Two robots tidy up a kitchen. They clean a spill, put away loose items, and fill the dishwasher [V10].   &  Hello Robot Stretch 3 & 33s \\

    \raisebox{-0.5\height}{\includegraphics[width=2.5cm, cfbox=black 0.6pt 0pt]{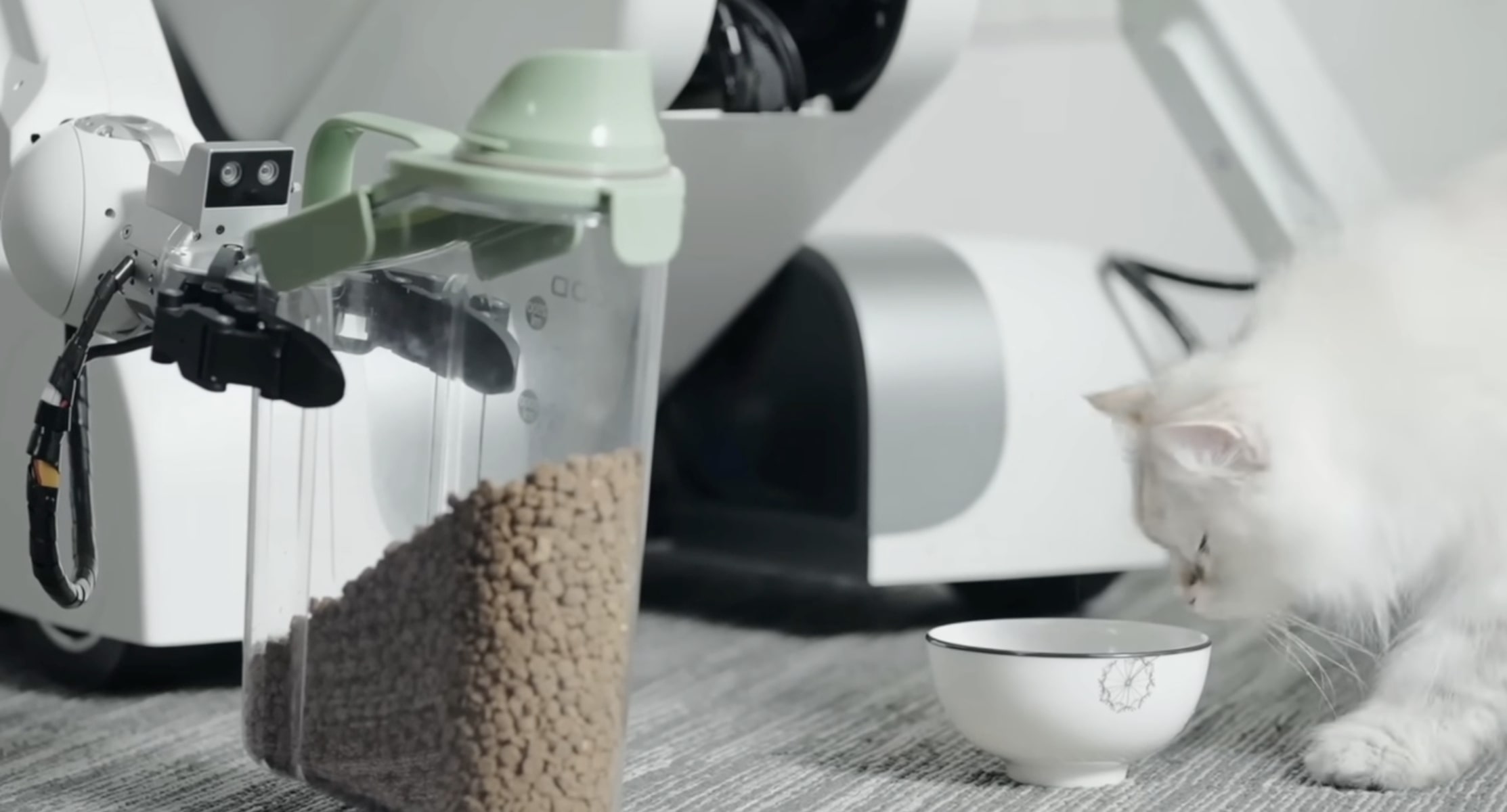}}  & A robot adds cat food to a bowl [V11]. & Astribot S1 & 10s \\

    \raisebox{-0.5\height}{\includegraphics[width=2.5cm, cfbox=black 0.6pt 0pt]{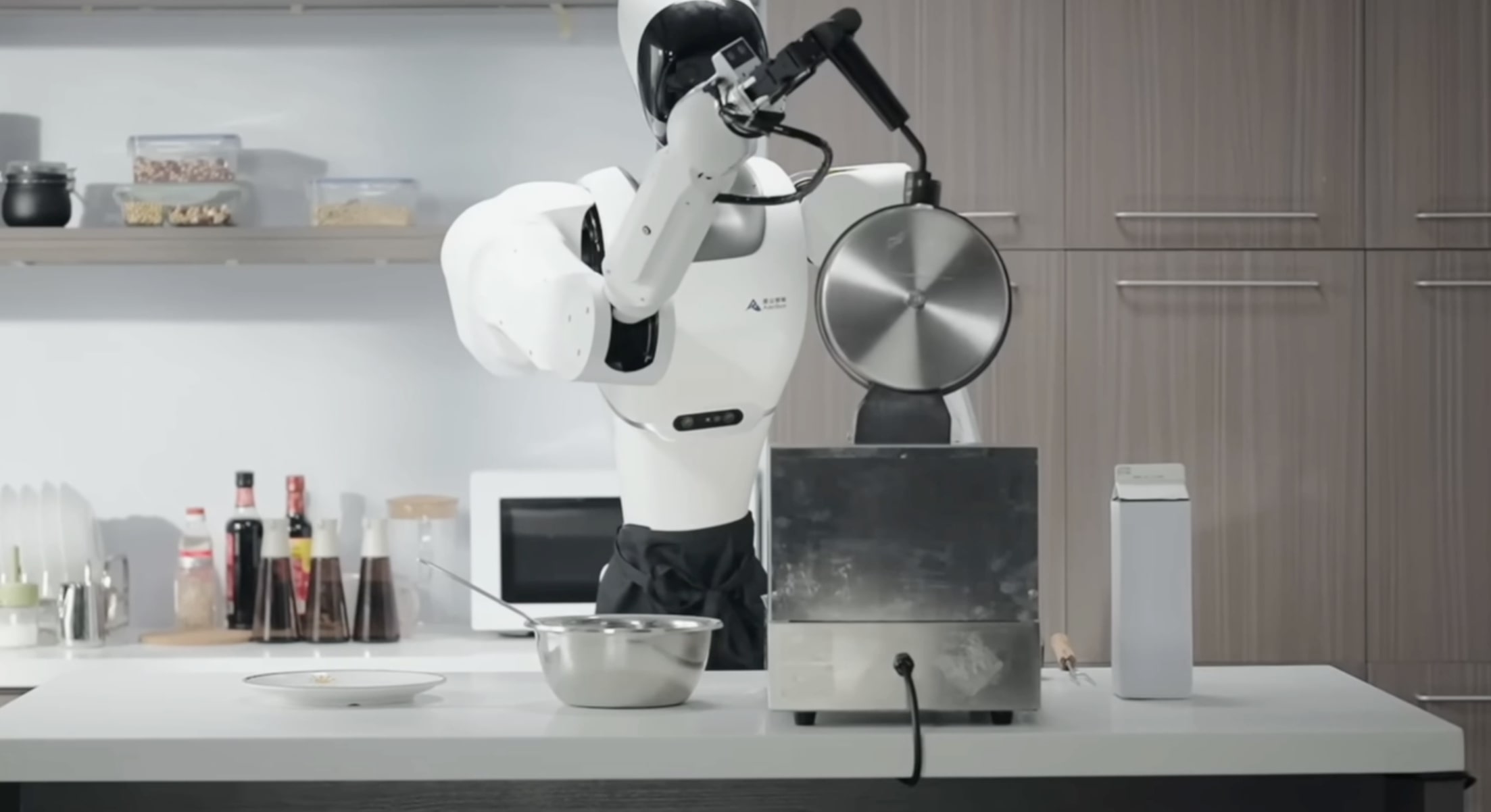}}  & The robot adds dough to a waffle iron and puts the finished waffle on a plate [V12]. & Astribot S1 & 36s \\

    \raisebox{-0.5\height}{\includegraphics[width=2.5cm, cfbox=black 0.6pt 0pt]{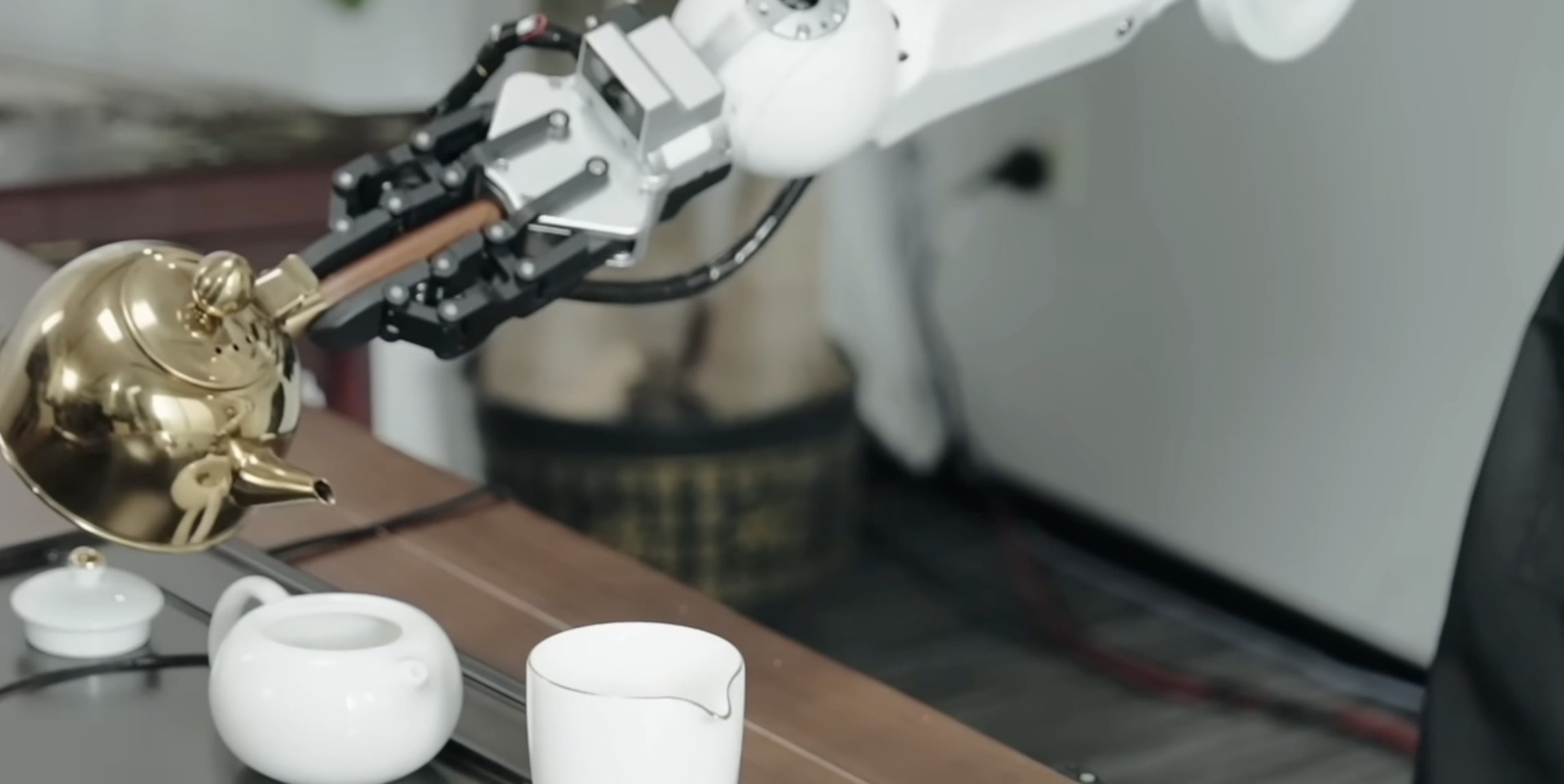}}  & The robot adds tea leaves and hot water to a pot. It then pours the tea and serves it to a person [V13]. & Astribot S1 & 27s \\

    \raisebox{-0.5\height}{\includegraphics[width=2.5cm, cfbox=black 0.6pt 0pt]{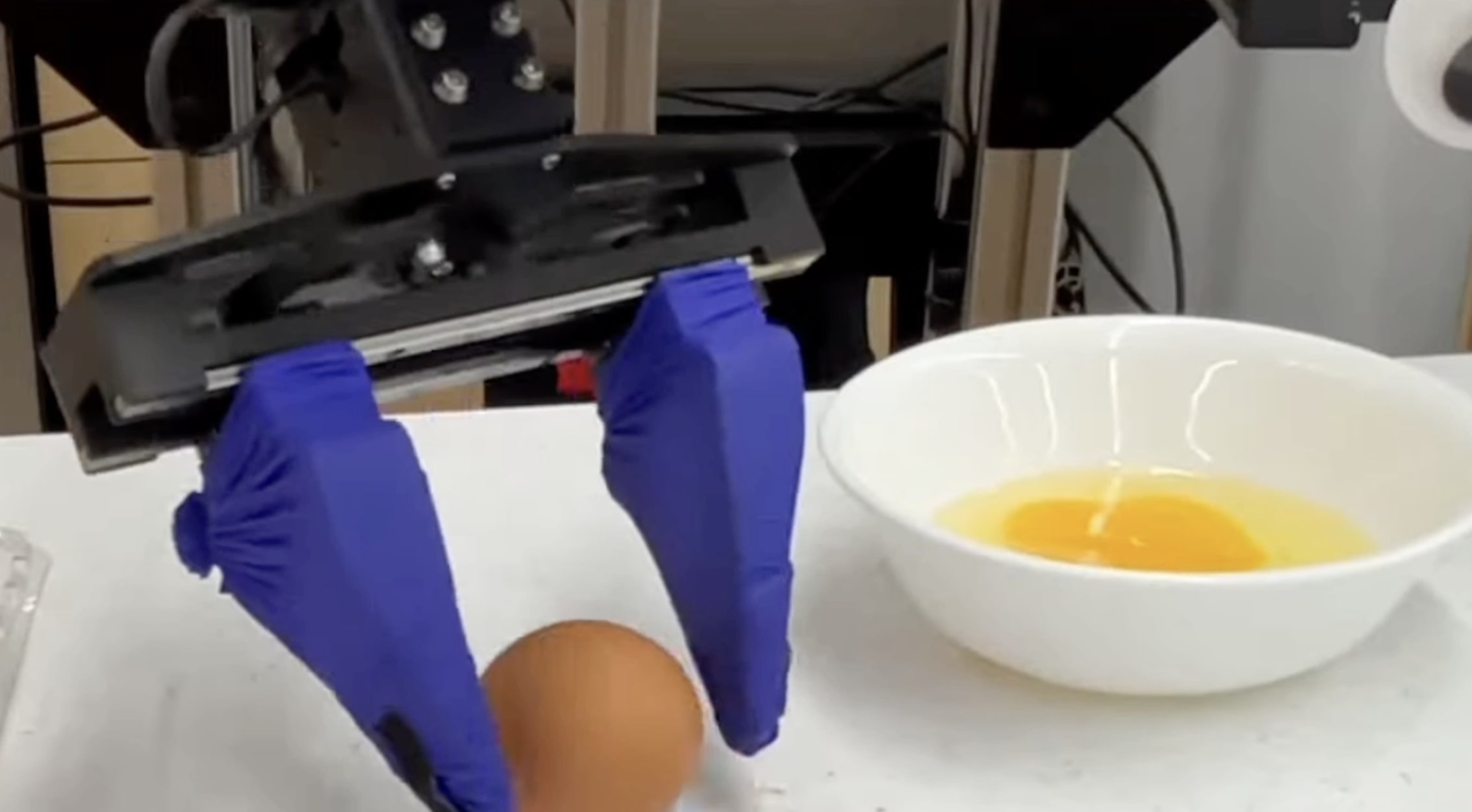}}  & The robot prepares two eggs and shrimp. It then mixes the two ingredients in a stir-fry [V14]. & Mobile ALOHA & 35s \\

    \raisebox{-0.5\height}{\includegraphics[width=2.5cm, cfbox=black 0.6pt 0pt]{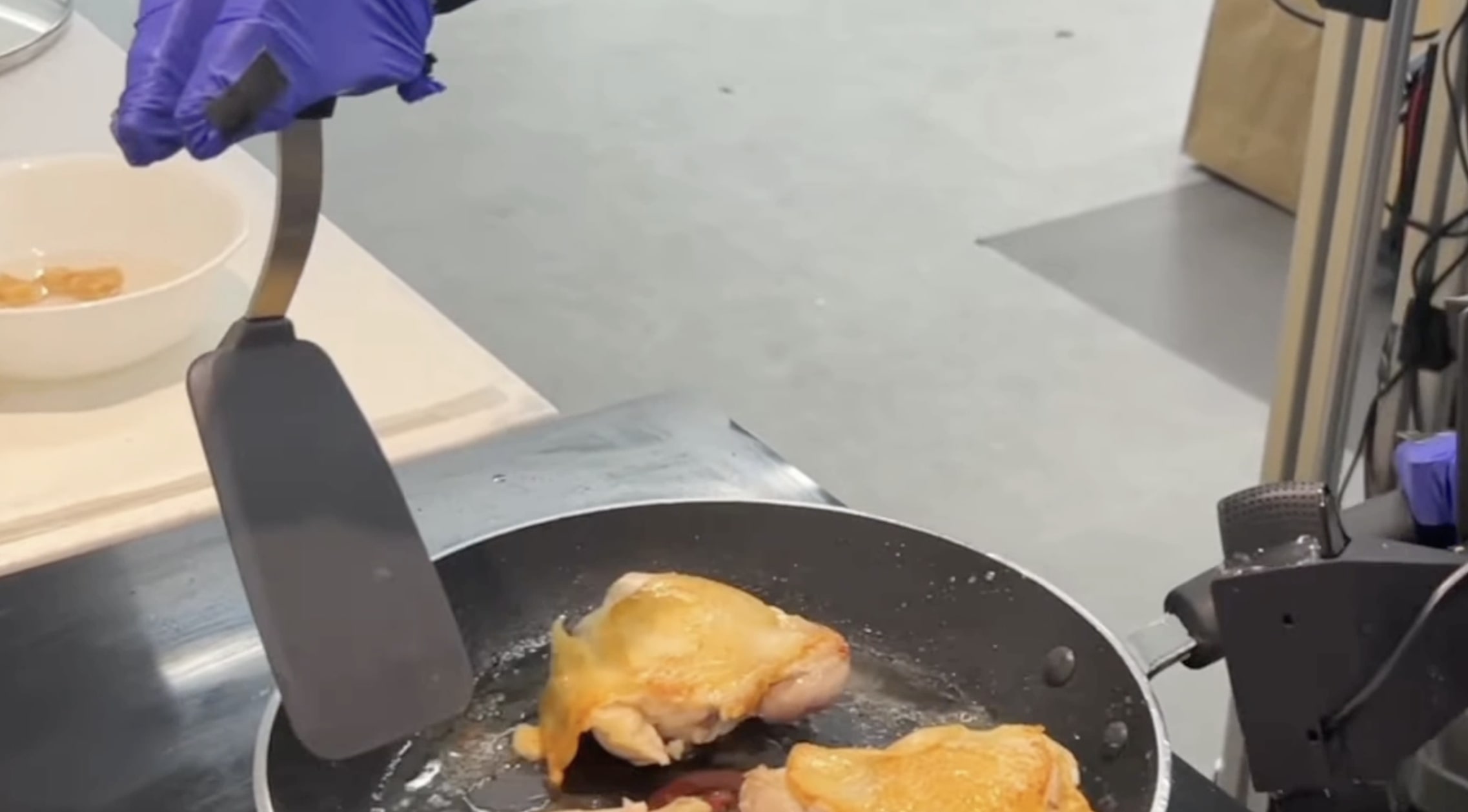}}  & The robot prepares some chicken meat in the pan, adding sauces and seasoning [V15]. & Mobile ALOHA & 41s \\

    \end{tabular}
        \end{adjustbox}
\end{table*}

\paragraph{Texts}
In addition to the videos, we wrote seven text reports of a robot's actions while its user was away. As with the videos, users were asked to write down the questions they might have for the robot in this scenario. The text scenarios contain a small number of failure reports, such as the robot breaking a plate while cleaning the dishes. An example text scenario is the following:
\begin{tcolorbox}
Imagine you have a robot at home that takes care of various household tasks. After returning from work, you check what your robot did today. It presents you with the following list.

1. I moved from the living room to the kitchen.

2. In the kitchen, I cleaned the dishes.

3. I moved from the kitchen to the living room.

4. In the living room, I picked up various items and put them back where they belong.

5. In the living room, I cleaned the windows.

6. I returned to my charging station. 
\end{tcolorbox}

A list of all text scenarios used can be found online (see Appendix \ref{app:online}).

\subsection{Participants}

\begin{figure}
    \Description[A visual summary of the participant statistics described in Subsection 3.3]{A visual summary of the participant statistics described in Subsection 3.3}
    \centering
    \includegraphics[width=1\linewidth]{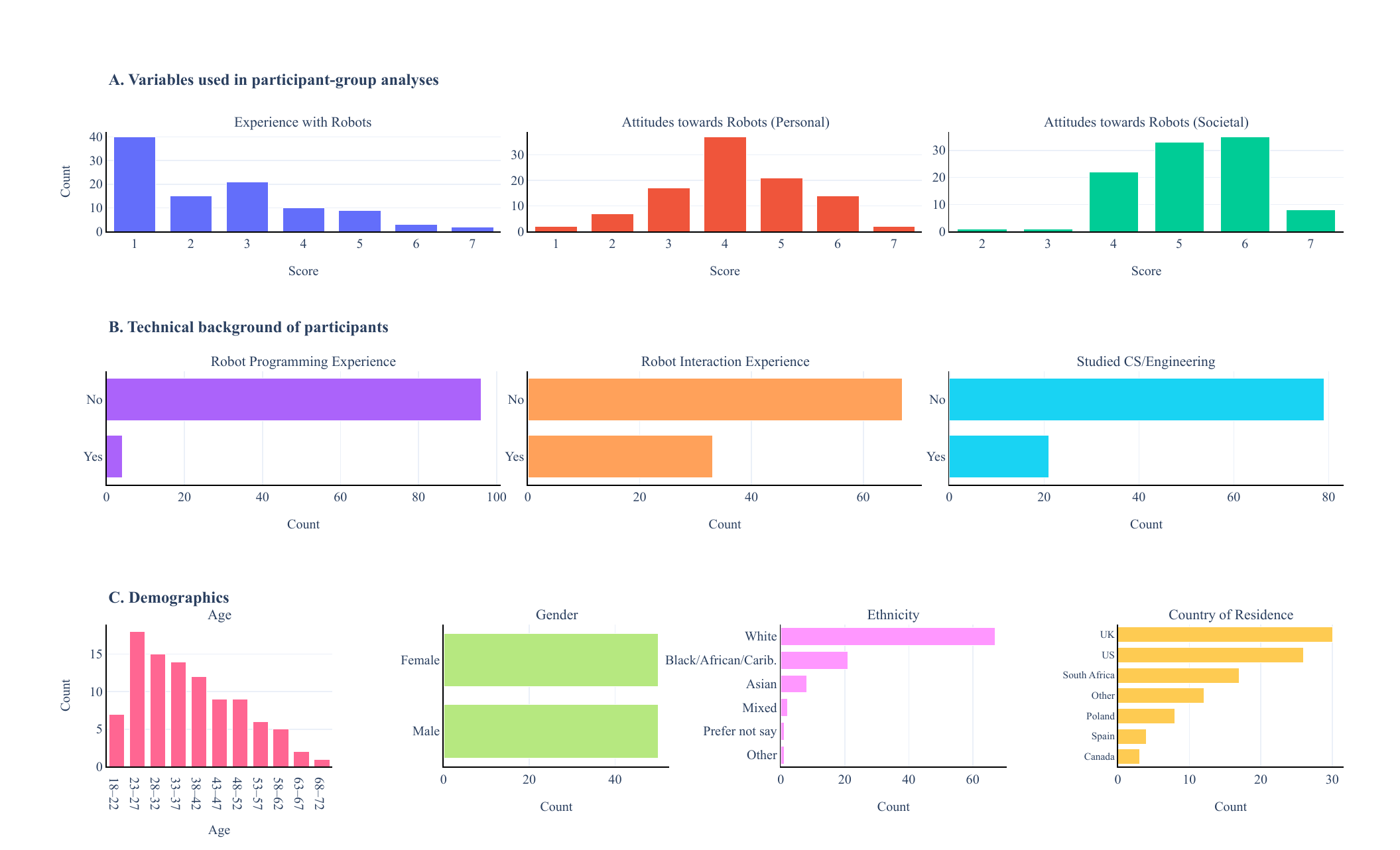}
\caption{Participant statistics grouped by variable type. 
A: variables used in the analysis of differences between participant groups. 
B: technical background of participants.
C: demographic variables. }
    \label{fig:participants}
\end{figure}

The final dataset consists of 1,893 questions from 100 participants. This excludes participants who failed any attention check or gave nonsense answers in the free-text input fields. On average, participants were paid an hourly wage of 9.4 GBP, which falls in Prolific's recommended range at the time of the study. 
Filters were set to include only participants who are fluent in English and who have a Prolific approval rate of 100\%, having participated in at least 5 studies. 
Of the accepted participants, 50 were female and 50 male. The age range was 18 to 73, with an average age of 32 years (SD=13). 
The respondents lived in the UK (30 participants), the USA (26), South Africa (17), Poland (8), Spain (4), Canada (3), and 12 other countries with two or fewer participants. Of these, 67 were white, 21 Black/African/Caribbean, 8 Asian (Indian, Pakistani, Bangladeshi, Chinese, any other Asian background), 2 mixed (two or more ethnic groups), 1 other (Arab or other), and 1 preferred not to say. Out of the 100 participants, 4 had programmed a robot before, 21 studied computer science, robotics, or engineering, and 33 reported having interacted with a robot. The participant statistics are visualized in Figure \ref{fig:participants}.

Ethical clearance was obtained from the King's College ethics committee (MRSP-24/25-50343).

\subsection{Methods for Data Analysis}

\paragraph{Category Coding (RQ1)} \label{sec:category_coding}
We coded all questions into categories in order to give structure to the dataset and make it into an actionable resource for future robotic system design. 
The creation of a new coding scheme was necessary as previous categorizations \cite{mcguinness2007categorization, raza2022wild} did not reflect the richness of questions we encountered in the context of users communicating with robots conducting household tasks. Moreover, manual annotation was necessary to ensure high quality, as automated methods, such as clustering, are difficult to interpret and struggle to capture fine-grained differences in user intent.  

Instead, the questions were coded by the first author of this paper in an inductive, bottom-up manner. No pre-defined (top-down) codes were used. We 
iteratively annotated the dataset of questions, initially with descriptive terms close to the content. We then iteratively merged related codes into higher-level themes. The final codes are structured in a two-level hierarchy.
After annotating all questions and defining all categories, the main annotator did a second pass over all questions to guarantee that the annotations agree with the final category definitions. 

To ensure understandability and reproducibility of the codes, we provide a codebook with definitions and examples of each code in Table \ref{tab:codebook}. Moreover, a second person re-annotated 96 questions, which were randomly sampled and balanced across categories, i.e., contained 8 samples per category. Comparing the two annotations, we find that the two annotators agreed on 79\% of the questions. Computing Cohen's Kappa, a chance-corrected coefficient of agreement, we obtain a value of .77 (CI = [.68, .86], $p<.0001$). This value corresponds to ``substantial agreement'' according to Landis et al. \cite{landisMeasurementObserverAgreement1977}. 
With three occurrences, the most common disagreement was between self/task-assessment and execution-details. As a consequence, we improved the definitions in the codebook to make the differences clearer.  
Other disagreements can be traced back to differences in the interpretation of the question. For example, ``Are you able to tell me if we're low on forks''	and ``If it can remember where it placed each item?'' can be interpreted as the user asking for whether the robot has the ability. However, it can also be interpreted as a way of asking for the availability of forks and the item locations. Such disagreements can only be resolved by asking the participant who provided the question. Future extensions of this dataset could benefit from participants providing example answers they would deem satisfying in order for the annotators to better understand their expectations.

\paragraph{Statistical Analysis --- Question Importance Ranking (RQ2)} \label{sec:method-importance}

To analyze the differences in assigned importance of the different question categories, we use a linear mixed-effects model \cite{lindstrom1988newton}.  It is suited to our dataset, where individual data points are non-independent, as multiple questions (of the same or different question categories) are rated by the same participant (within-participant). Simply averaging each participant's ratings would mean a loss of valuable information, which is why we use linear mixed-effects models.
Therefore, we model the participant identifier as well as the stimulus identifier as random effects in our mixed-effects model. We use a random intercepts model, meaning the observations per participant/stimulus are moved by a dedicated intercept.
The fixed effect is the question category. The outcome variable is the importance score. A linear mixed-effects model, as described, can be written down as: 
\[
importance \sim category + (1 | participant) + (1 | stimulus)
\]
The model thus accounts for within-participant and within-stimulus correlation, which in turn provides us with valid standard errors. The model was fit by restricted maximum likelihood.
We correct for multiple comparisons via the  Holm–Bonferroni correction \cite{holm1979simple}. This ensures the family-wise error rate, i.e., the probability of making one or more false discoveries, stays below $\alpha<.05$.

\paragraph{Statistical Analysis --- Influence of Attitudes towards Robots and Robotics Experience on Question Importance (RQ3)}

Given the varying explanation needs of different users shown in XAI research (see Section \ref{sec:rel-user-dif}), we decided to explore the relationship between user characteristics (their attitudes towards robots \cite{koverola2022general} and robotics experience) and their question importance scores for different categories.

For this analysis, we compute correlations between user-related variables. We use Spearman's rank correlation coefficient $\rho$ \cite{spearman1904proof} to measure the monotonic relationship between variables. It ranges from -1 to +1, with a positive/negative value indicating that the variables tend to increase/decrease together.
Again, we correct for multiple testing via the Holm–Bonferroni correction \cite{holm1979simple}.
Lastly, we run a multiple linear regression \cite{montgomery1982introduction} to gain further insight into how user characteristics relate to how much importance users place on the robot's ability to answer their questions. To do so, we use attitudes (societal/personal-level) and robotics experience as predictor variables, and question importance as the outcome variable. Given that the attitude scores are averages of multiple Likert scale items, we can treat them as continuous variables. While the robot experience variable represents scores on a 1--7 scale, it is accepted practice to use such scores as a numeric variable in statistical models like linear regression \cite{norman2010likert}.

Statistical models and tests are run via the Python libraries SciPy \cite{2020SciPy} and statsmodels \cite{seabold2010statsmodels}.

\section{Results --- A Dataset of Questions for a Robot}
The results of the study, including participant answers and analysis code, are made available as a public dataset\footnote{https://github.com/lwachowiak/xai-questions-dataset}. 

Overall, we collected 2,037 questions. From these, we excluded 144. Most of the excluded questions were commands in the form of a question, for example, ``Could you clean the fridge?'' The remaining excluded questions are hard-to-interpret phrases with missing words or unclear meaning, as well as very rare cases of users writing non-serious joke questions. On average, each participant provided 18.93 (SD=2.82) questions to the dataset. The average question length is 9.73 words (SD=4.16).
The questions asked in the dataset show linguistic diversity. 
Most users speak to the robot directly (67\%), for example, ``What kind of ingredients did you put in the salad?'' In contrast, other users refer to the robot (or sometimes the environment) in third-person (30\%), for example, ``Why did it flip the waffle at the end?'' In rare cases (2\%), the questions are asked from a 1st person perspective, for example, ``What items do we have to make a meal?''
Some of the questions can only be made sense of in the concrete context of the text or video scenario. For example, ``What about the bathroom floors?'' only makes sense in the context of the robot having reported mopping the kitchen floor.
In terms of extracted tense, most questions are written in the past (50\%) or present (48\%), and a small subset in a future tense (2\%). However, the (auxiliary) verb tense does not necessarily need to indicate the future for the question to be about potential future actions, as in ``Are you considering adding any other ingredients other than the ones on the table?''


\begingroup
\small
\renewcommand{\arraystretch}{1.2}  
\begin{longtable}{
    @{}
    >{\justifying\setlength{\parindent}{0pt}\setlength{\parskip}{0pt}\arraybackslash}p{0.20\textwidth}|
    >{\justifying\setlength{\parindent}{0pt}\setlength{\parskip}{0pt}\arraybackslash}p{0.40\textwidth}|
    >{\justifying\setlength{\parindent}{0pt}\setlength{\parskip}{0pt}\arraybackslash}p{0.33\textwidth}
    @{}
}
\caption{Codebook of question categories and subcategories. For each main category, the three most frequent subcategories are defined and illustrated with examples. Less frequent subcategories are summarized as ``other''. Categories are ordered by descending frequency.}
\label{tab:codebook}\\

\textbf{Code/Category} & \textbf{Definition} & \textbf{Example from the Dataset} \\
\toprule

\textbf{execution-details} & & \\

what-actions & Qs about whether the robot executed a specific action & Did you power off the washing machine?\\

what-objects & Qs about environment objects that were relevant during the task  &  What kind of ingredients did you put in the salad?\\
        
how-was-it-done & Qs about how exactly the robot did the task &  What temperature did you cook the food at? \\ 

other & Qs about further details of the task execution, e.g., where something was put, what goal was pursued, or in what order actions were executed & \\

\midrule

\textbf{what-abilities} & & \\

 extra-ability & Qs about whether the robot can execute an additional skill mentioned in the question & Can you wash the dishes for me too?\\

 ability-modification & Qs about whether the same skill can be done in a different way & Can you do that faster?\\
        
 ability-list & Qs that prompt the robot to list its abilities either generally or of a specific subcategory  &  What other kind of chores can you help me with? \\ 

other & Qs about the limits of the robot's abilities and whether it can learn certain skills & \\

\midrule

\textbf{self/task-assessment} & & \\

 difficulty & Qs about whether the task was difficult or easy for the robot & How easy did you find closing the cupboard?\\

 did-you-ensure & Qs about whether the robot ensured it is doing things correctly/safely/etc. & Did you check the smoke alarm was functional before you started cooking?\\
        
 correctness & Qs about whether the robot did everything correctly or made any errors & Is it cooked enough to be safe to eat? \\ 

other & Qs about the robot's general performance, confidence-levels, task success, and whether any problems arose & \\

\midrule

\textbf{why} & & \\

 why-did-you & Qs about why the robot did something (without the question explicitly proposing a contrast/alternative) & Why did you put the eggs in the fridge?\\

 why-not-alternative & Qs about why the robot did not take an alternative course of action, explicitly mentioned in the question & Why did you throw away the milk immediately instead of asking me first?\\
        
 why-issues & Qs about why the robot failed or was unable to do a task  &  Why did you bump into the bed? \\ 

other & Qs about why the robot did not do an additional task mentioned in the question or why it followed a specific action order & \\

\midrule

\textbf{environment-state} & & \\

object-properties & Questions (Qs) about the state/properties of the external environment & Did the dishes look rinsed when you put them in the dishwasher?  \\

availability & Qs about what is available, if something is available, or the quantity that is available & What items do we have to make a meal? \\
        
people/pets & Qs about people/pets the robot interacted with & Was the cat afraid of you? \\ 

other & Qs about how the user can/should interact with the environment & \\

\midrule

\textbf{time} & & \\

how-long & Qs about time lengths (task, actions, expiry dates, etc.)  & How long did it take you to tidy the room? \\

at-what-time & Qs about when something happened/is going to happen & At what time did you receive the mail? \\
        
frequency & Qs about how frequently something occurs & How many times a day do you feed the cat? \\ 

task-count-within-time & Qs about the number of tasks the robot can accomplish in a set time frame & How many clothing items can you fold in an hour? \\ 

other & Qs about frequencies of actions/tasks, and about comparing task lengths  &  \\ 

\midrule

\textbf{how-did-you-know} & & \\

 goal-properties & Qs about how the robot knows specific desiderata of its goal/task & How did the robot know the right time to feed the cat?\\
        
 goal-achieved & Qs about how the robot knows it achieved a goal or subgoal  &  How does the robot know when the food is ready to be served? \\ 

  object-properties & Qs about how the robot knows an object has a specific property &  How do you know the difference between clean and dirty dishes?\\

other & Qs about how the robot knows various other things, such as the right steps to take, which task to prioritize, or which objects are relevant to a goal   & \\

\midrule

\textbf{mental-state} & & \\

check-understanding/beliefs& Qs about the robot's understanding of the world, tasks, and instructions & What tasks do you believe is part of cleaning the room? \\

emotions & Qs about the robot's ``emotions'' (joy, boredom, etc.) & Did you enjoy the task? \\
        
opinion/likes & Qs about the robot's opinions and likes & What's your perfect lunch box? \\ 

other & Qs about other mentalistic concepts, e.g., the robot's ideas, inner thoughts, or desires & \\

\midrule

\textbf{potential-issues} & & \\

 what-if & Qs about hypothetical scenarios involving challenges, errors, and changes in external circumstances  & What happens if you drop a piece of clothing? \\

 how-to-ensure-correctness & Qs about how the robot ensures that a task is executed correctly & How do you detect and correct mistakes during folding? \\
        
does-it-happen-that & Qs about whether specific potential issues can arise & Do you ever mistakenly misplace things?\\

\midrule

\textbf{how-did-you-decide} & & \\

 goal-properties & Qs about how decisions about goal details, like location, arrangement, or amount, were made & How did you decide where each toy should go when you clean up? \\

 goal-objects & Qs about which items were part of the goal(-configuration) &  How did you pick the lunchbox items? \\
        
 task-priority & Qs about how the priority and order of tasks were determined & How do you decide which household tasks should be done first? \\ 

other & Qs about how decisions were made, including regarding the actions taken, when to stop, or object properties determined  & \\

\midrule

\textbf{future-actions} & & \\

what-needs-to-be-done & Qs about which tasks in the house need to be done now or in the future & Does the rug need to be vacuumed? \\

what-next & Qs about what actions/tasks the robot will do next & Will you dry the dishes after washing them? \\
        
assistance-need & Qs about whether the robot requires help from the user & Do you have anything that I can move out of the way so you can work more efficiently? \\ 

other & Qs about what external requirements need to be fulfilled for the robot to be able to do the task; and about how it can improve in the future& \\

\midrule

\textbf{technical-details} & & \\

how-does-it-work & Qs about technical details of how the robot functions & How do you see the objects around you? \\

who-has-control & Qs about autonomy and who can give commands & Are other humans able to give you tasks? \\ 

battery-status & Qs about battery levels & Do you need to be charged? \\

other & Qs about hardware and software specifics &  \\ 

\bottomrule
\end{longtable}
\endgroup

\subsection{Question Categories}

\begin{figure*}[!htbp]
    \Description[A Sankey diagram showing the distributions of questions among the different categories and subcategories.]{A Sankey diagram showing the distributions of questions among the different categories and subcategories. Most questions belong to the category execution-details (21.4\%), followed by what-abilities (12.6\%), self/task-assessment (10.7\%), why (9.8\%), environment-state (8.1\%), time (6.5\%), how-did-you-know (6.4\%), mental-state (6.2\%), potential issues (6.1\%), how-did-you-decide (5.4\%), future-actions (4.3\%), and technical-state (2.7\%)}
    \centering
    \includegraphics[width=\linewidth]{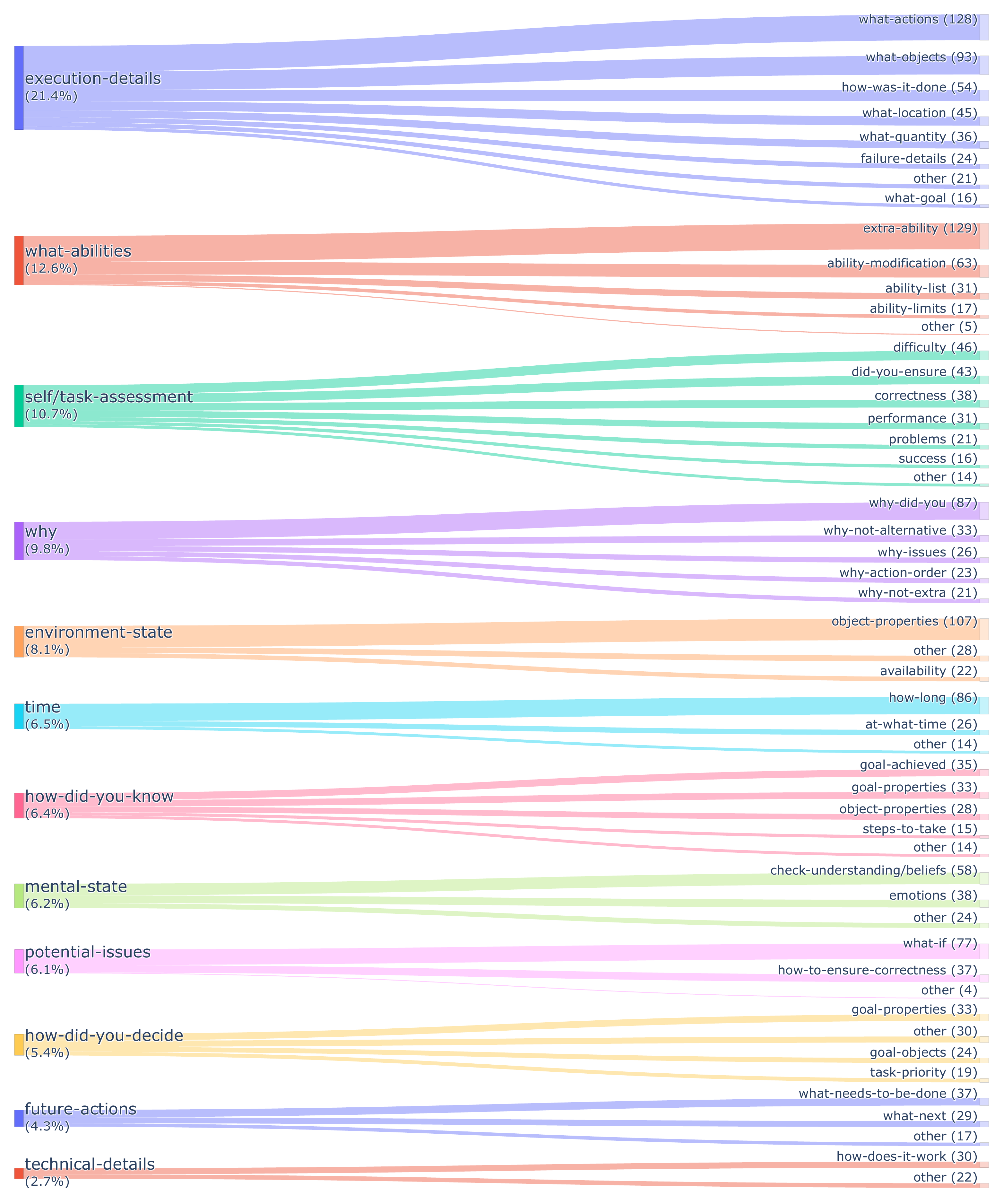}
    \caption{Hierarchical categorization of user questions for the robot (see Table \ref{tab:codebook} for category definitions).}
    \label{fig:category-graph}
\end{figure*}

Our dataset shows a high variety of questions and themes. Figure \ref{fig:category-graph} gives an overview of which categories and subcategories are the most prevalent. Table \ref{tab:codebook} provides the definitions of the main categories as well as each of their three most common subcategories. While some question (sub)categories are very prevalent, for example, questions of the form ``can you do <x>'', there is a long tail of rare question categories. This variety of questions and the long tail of rare question types show how difficult it is to develop question-answering components in robotics that can reliably work in all situations. 

\paragraph{Execution Details.} The most common in the dataset are questions in the category execution-details, which deal with the specifics of what happened during the robot's task execution. These are often what-questions, querying, for instance, what the robot did, what objects it used, or what goal it was trying to achieve.

\paragraph{Abilities.} The second largest category is questions that ask the robot about its capabilities. These questions ask about specific additional abilities, modified versions of what the robot already did, or all the abilities the robot has in a specific area. Less commonly, users also asked about the limits of a robot's abilities, for instance, what would be too heavy or too far away.

\paragraph{Self/Task-Assessment.} In this category, people ask the robot to assess how the task went, for example, if there were any problems, if the robot was successful, or if the task was difficult.  

\paragraph{Why.} Participants also provided many why-questions, a category of questions studied in detail throughout the XAI literature. Questions in this category ask the robot to give reasons for its actions. These questions are usually action-focused. Sometimes, participants provide an explicit contrast, so the robot has to answer why it chose one course of action over another. An example of a contrastive why-question is ``Why did you decide to pick an orange and not any other fruit?'' Such an explicit contrast is given in 28\% of the why-questions. However, even in the case of questions without explicit contrast, people would expect contrastive explanations according to XAI literature \cite{miller2019explanation}. For instance, when asked ``Why did you choose the orange?'', the robot could automatically infer the contrast by identifying what other fruits were on the table as alternatives. 
Furthermore, why-questions are asked for varying reasons. For instance, in the question
``Why did you put the tools away in the places that you did so next time I can find them?'', the participant explicitly mentions that a causal explanation would help them to better predict the robot's actions in the future. 

\paragraph{Environment State.} A similar category in the sense that it mostly deals with simple facts is the environment-state category. In contrast to the execution-details category, these questions are not about what the robot did but are about the state of the external environment, e.g., ``Are there any foods that have expired?''

\paragraph{Time.} Beyond the question of whether a robot can achieve a task, people were also interested in how long the robot needs to execute a task. In addition, this category includes questions about when something happened or will happen. There is also a small set of questions about the frequency of robot actions, as well as how many tasks the robot can achieve in a fixed amount of time. 

\paragraph{How Did You Know.}
While similar to the categories of why- and how-did-you-decide-questions, this category focuses less on the decision-making process. Instead, it emphasizes the robot's knowledge and where this knowledge comes from. Examples of knowledge sources can be the robot's perception, learning, or prior information. Common subcategories are concerned with properties of the goal and the environment, with questions such as ``How did the robot know the right time to feed the cat?'' (goal property) or ``How do you know if an item is dishwasher friendly?'' (object property). 

\paragraph{Mental State.} This category of questions targets the robot's mentalistic concepts. The largest subcategory contains questions that probe the robot's understanding and beliefs, e.g. ``What tasks do you believe are part of cleaning the room?'' The second largest subcategory contains questions that ask about the robot's emotions, such as whether it found the task enjoyable, boring, or frustrating. 
Questions from other subcategories include asking the robot about its opinions, ideas, preferences, or desires. 
However, whether a robot should ascribe some of these mental concepts to itself in answers is a controversial topic. For instance, in a survey by the UK AI Security Institute, 61\% of participants thought that it is wrong for a chatbot to express emotions, while only 19\% thought it was acceptable \cite{aisi2024-should-ai-behave}. With mental concepts such as preferences and beliefs, the same participants were more undecided; however, they still leaned towards it being unacceptable.   

\paragraph{Potential Issues.} Questions in this category are about what potential issues exist and how the robot would handle them. Such issues include both robot errors as well as environment challenges. Questions also pertain to how the robot can ensure it executes tasks correctly so as not to run into such issues.

\paragraph{How Did You Decide.}
With questions of this category, users ask the robot about how it made a decision. Similar to the questions in the \textit{why} category, questions here may be answered by referring to causes and alternatives. However, depending on the user intent, answers not referring to causes are also possible. Specifically, ``how did you decide...'' can be interpreted as placing more emphasis on the underlying decision-making procedure (corresponding to Marr's algorithmic level \cite{marr2010vision}) rather than on the causes. 
For instance, the question ``How did you decide how much water to use on the plants?'' could be answered by saying ``I measured the soil moisture and compared it to the recommended target'', thus emphasizing the algorithm employed.
Common subcategories concern the decision-making process behind the details of the goal, such as which objects to involve or goal-properties like where to place something.

\paragraph{Future Actions.} This category of questions deals with what the robot plans to do next. The most common subcategories include questions on the robot's next actions, whether it needs assistance, and what needs to be done.

\paragraph{Technical Details.} Questions in this category are concerned with mechanistic explanations of how the robot functions, e.g., what algorithms or models it uses or how its visual perception works. Questions in this category also touch upon the robot's settings, how to change them, and who can control the robot. Lastly, the category contains questions about the robot's hardware and battery functionality. 


\subsection{Question Importance Scores} \label{sec:res-imp}

\begin{figure}[!htbp]
    \Description[A graph showing the estimated marginal means and 95\% confidence intervals for the question importance score per question category. Additionally, the number of samples per category and pairwise statistically significant differences are indicated.]{A graph showing the estimated marginal means and 95\% confidence intervals for the question importance score per question category. Additionally, the number of samples per category and pairwise statistically significant differences are indicated. The highest importance score is found for the category potential-issues (4.011), followed by how-did-you-know (3.952), self/task-assessment (3.861), how-did-you-decide (3.848), execution-details (3.810), technical-details (3.744), environment-state (3.697), future-actions (3.661), what-abilities (3.639), time(3.630), why (3.582), and mental-state (3.534). A list of the pairwise differences that are statistically significant can be found in Section \ref{sec:res-imp}.}
    \centering
    \includegraphics[width=\linewidth]{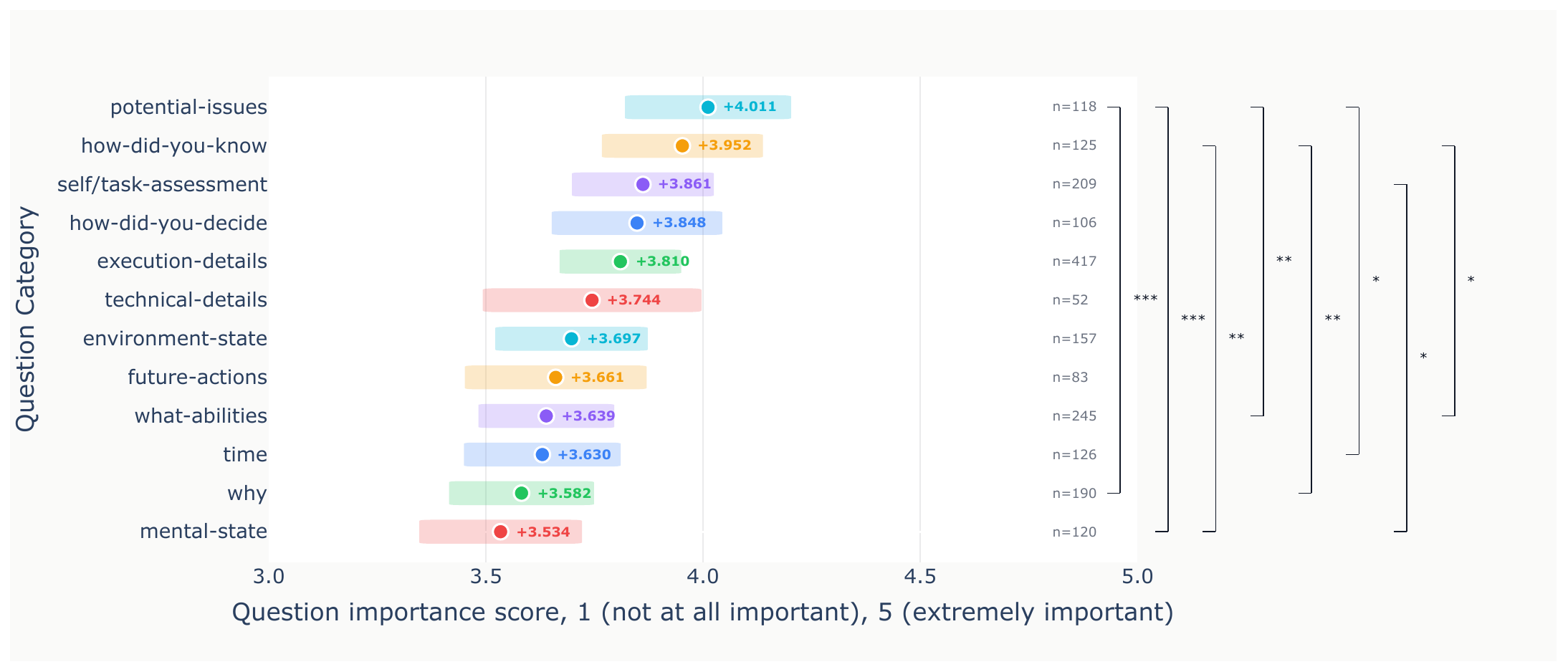}
    \caption{Estimated marginal means (95\% CIs) of the importance scores per question category --- obtained via a linear mixed-effects model. Number of samples per category (n) shown on the right, and statistical significance indicated by: *** = $p<.001$, ** = $p<.01$, * = $p<.05$}
    \label{fig:imp-graph}
\end{figure}

As explained in Section \ref{sec:method-importance}, we modeled the question importance via a linear mixed-effects model. The question category (e.g., why-questions) is the fixed effect, while the participant ID and the stimulus ID were modeled as random intercepts. 
As a result, we find the participant-level standard deviation (0.58) to exceed the stimulus-level standard deviation  (0.26), indicating greater heterogeneity across participants than across stimuli. This means that a participant's random intercept, which models their baseline importance, typically varies across individuals by $\pm0.58$ points on the 1--5 scale. 

Figure \ref{fig:imp-graph} illustrates the resulting estimated marginal means (EMMs) per category, as well as the respective 95\% confidence intervals (CIs). After accounting for participant- and stimulus-level variability, we find the category of questions about potential issues to be rated most important (EMM = 4.01, 95\% CI = [3.82, 4.20], n = 118). These are questions about what the robot would do when facing a (difficult) variation of the task, and how it can ensure it acts correctly and avoids issues.  
Questions rated the least important are in the why-questions (3.58, [3.42, 3.75], n = 190) and mental-state questions (3.53, [3.35, 3.72], n = 120) categories. 
We find multiple statistically significant differences between the importance ratings of the categories. After correcting for multiple comparisons, we find $p<.001$ with:
\begin{itemize}
    \item ($\uparrow$) potential-issues vs. ($\downarrow$) why (difference in EMMs = 0.43, $p=.0007$)
    \item potential-issues vs. mental-state (0.48, $p=.0009$)
\end{itemize}

We find $p<.01$ with:
\begin{itemize}
    \item how-did-you-know vs. mental-state (0.42, $p=.0051$)
    \item potential-issues vs. what-abilities (0.37, $p=.0058$)
    \item how-did-you-know vs. why	(0.37, $p=.0066$)
\end{itemize}

Lastly, we find $p<.05$ with:
\begin{itemize}
    \item potential-issues vs. time	(0.38, $p=.0239$)
    \item self/task-assessment vs. mental-state (0.33, $p=.0388$)
    \item how-did-you-know vs. what-abilities (0.31, $p=.0423$)
\end{itemize}

\subsection{Differences between Participant Groups}

\begin{figure}[!htbp]
    \Description[Left, a correlation matrix between participant-related variables (attitudes (personal-level), attitudes (societal-level), question importance, robotics experience). Right, a graph showing the linear regression coefficients for the same variables when predicting the user's average question importance rating.]{Left, a correlation matrix between participant-related variables (attitudes (personal-level), attitudes (societal-level), question importance, robotics experience). Right, a graph showing the linear regression coefficients for the same variables when predicting the user's average question importance rating --- with attitudes (personal-level) being the strongest predictor (0.143), attitudes (societal) the second (0.057), and robotics experience the last (0.054).} 
  \centering
  \begin{subfigure}{0.48\textwidth}
    \centering
    \includegraphics[width=\linewidth]{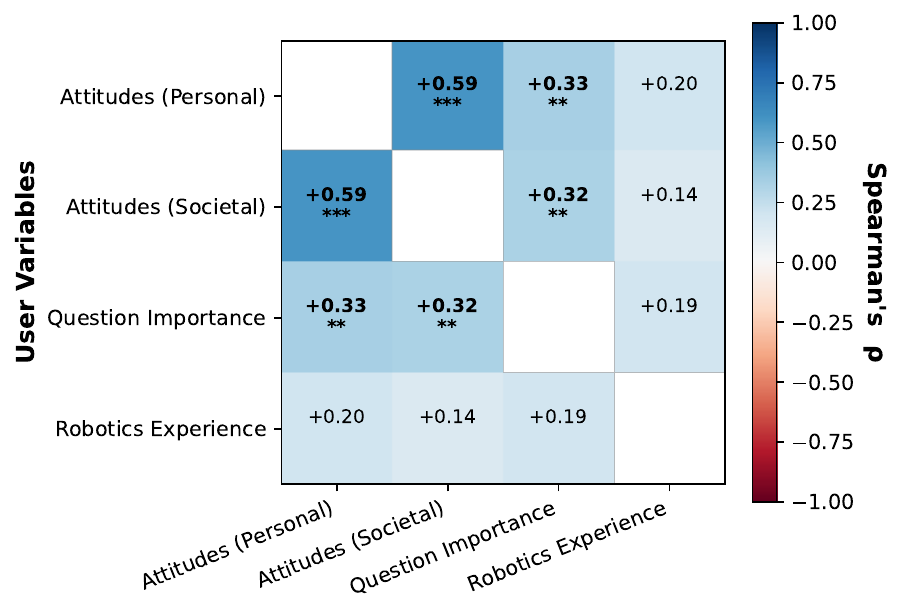}
    \caption{Correlations between a user's robot experience, avg. question importance, and personal/societal-level attitudes towards robots. *** = $p<0.001$, ** = $p<0.01$, * = $p<0.05$}
    \label{fig:corr-heatmap}
  \end{subfigure}\hfill
  \begin{subfigure}{0.48\textwidth}
    \centering
    \includegraphics[width=\linewidth]{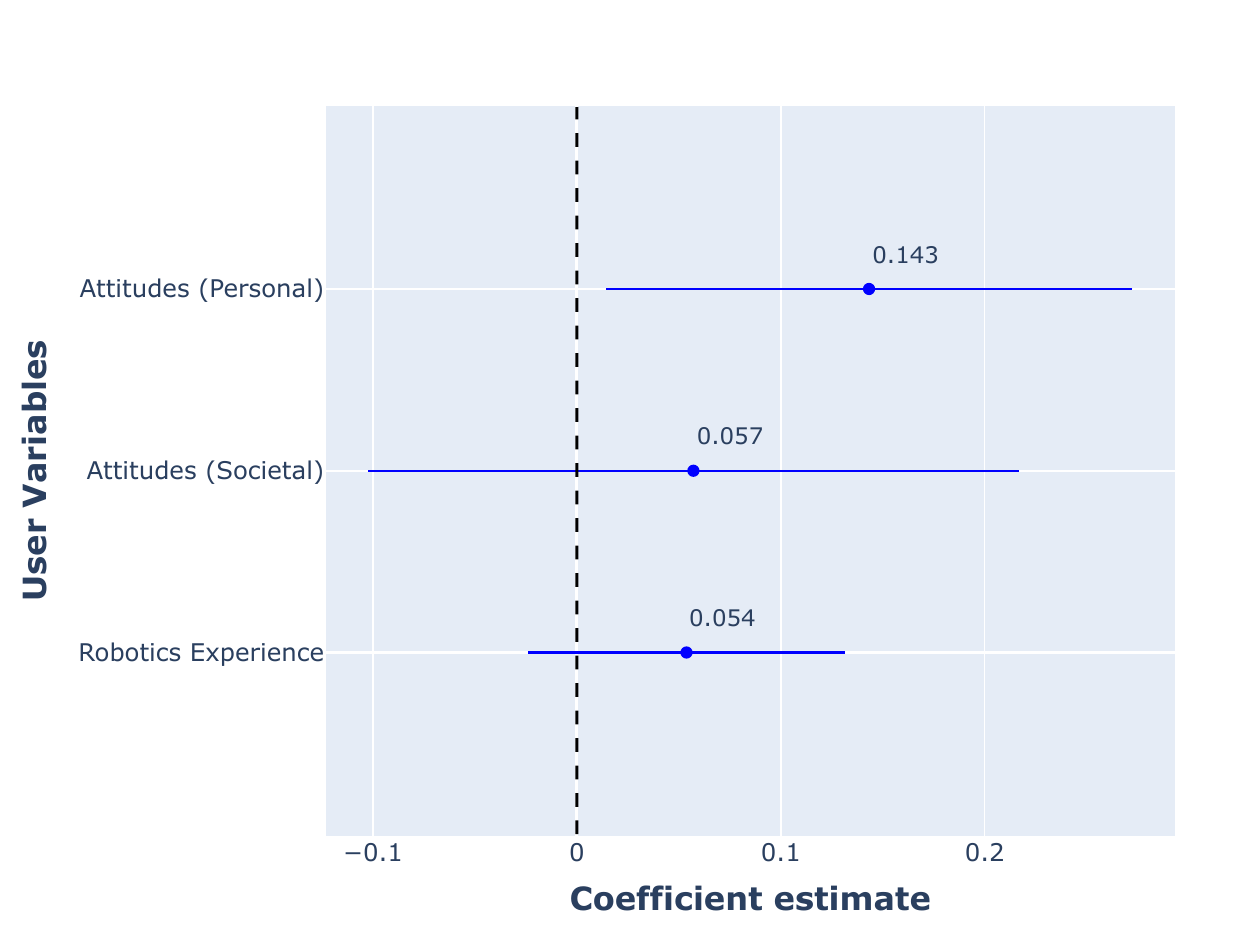}
    \caption{Linear regression coefficients (95\% CIs) with robot experience and personal/societal-level attitudes towards robots as a predictor of a user's average question importance rating.}
    \label{fig:linear-reg}
  \end{subfigure}
  \caption{Users' rating of how important they think it is that a robot can answer their proposed questions is related to their general attitudes towards robots as measured by the GAToRS questionnaire.}
  \label{fig:two-pdfs}
\end{figure}

\begin{figure}[!htbp]
    \Description[A bar chart showing the differences in question type distribution based on how participants rated their robot experience]{A bar chart showing the differences in question type distribution based on how participants rated their robot experience. The differences are most pronounced with the categories environment-state and execution-details. Both are more common with users with low robot experience. }
    \centering
    \includegraphics[width=0.9\linewidth]{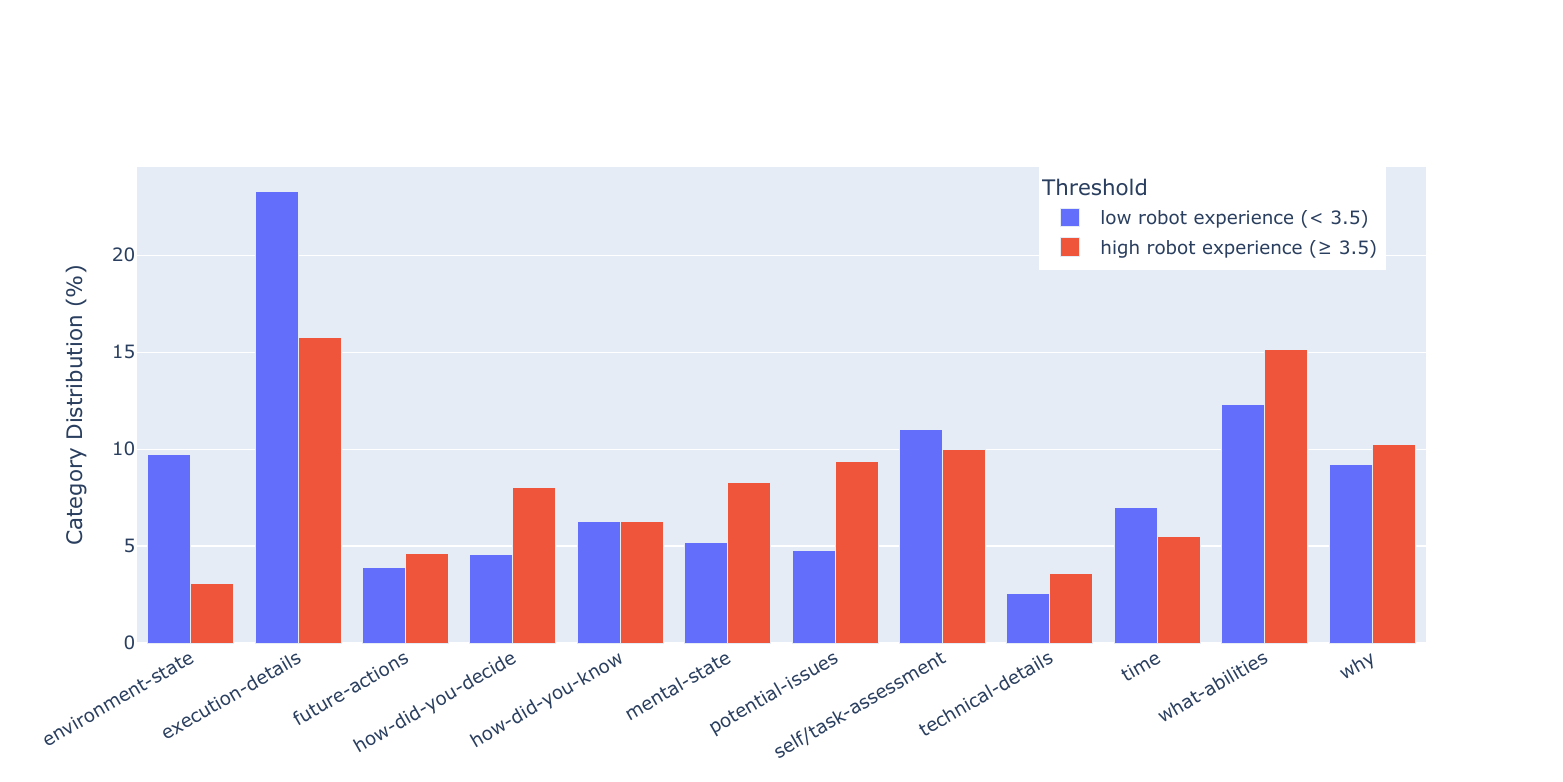}
    \caption{Difference in question type distribution based on robot experience. Robot experience is based on self-report scores on a Likert scale 1--7. There are 76 participants with a score between 1 and 3, and 24 participants with a score between 4 and 7.}
    \label{fig:robot_exp_diff}
\end{figure}

To aid future research on adapting explanation capabilities to individual users, we examined the impact of a user's robotics experience level and attitudes towards robots on what questions the user asks and how important they think it is for a robot to be able to answer their question.

\paragraph{Correlations with Question Importance Scores.}
We find that users' question importance rating has statistically significant correlations with users' personal-level ($\rho=.33, p = .0037$) and societal-level ($\rho=.32, p = 0.0047$) attitudes towards robots. Users' robotics experience is less strongly correlated ($\rho=.19, p=.154$) and not statistically significant. A full correlation heatmap is visualized in Figure \ref{fig:corr-heatmap}.

\paragraph{Linear Regression for Predicting the Importance Score.}
We examined whether users’ averaged question importance ratings could be predicted from their attitudes towards robots (personal/societal-level), and robot experience. The multiple linear regression model was statistically significant overall, 
$F(3, 96) = 5.26, p = .002$, explaining 14.1\% of the variance in the importance score. The resulting linear regression equation is:
\[
Avg.\ Importance\ Rating = 2.62 + 0.14(Personal\mbox{-}Level\ Attitudes) + 0.06(Societal\mbox{-}Level\ Attitudes) + 0.05(Robot\ Exp.)
\]

The linear regression shows personal-level attitudes towards robots to be the strongest predictor with a coefficient of 0.14 ($CI=[0.014,0.272]$), indicating that the predicted importance rating increased by $0.14$ points with every point increase in the personal-level attitudes scale. Societal attitudes are less important than in the correlation analysis, likely due to the high correlation between the two attitude scales ($\rho=0.59, p<.001$). Coefficients and CIs are visualized in Figure \ref{fig:linear-reg}.

\paragraph{Differences in Question Categories.} Figure \ref{fig:robot_exp_diff} shows how users with different levels of robot experience ask different types of questions. To create the graph, participants' contributions to each category were normalized, so a participant's contributions across all categories add up to one. This normalization was necessary as some participants contributed slightly more questions than others. 
Participants with a low robot experience score (1--3, n = 76) asked execution-details questions 23.3\% of the time. In contrast, participants with a high robot experience score (4--7, n = 24) asked execution-details questions only 15.7\% of the time (difference = -7.6\%). A similar contrast exists with environment-state questions, which were asked 9.8\% of the time by participants with low robot experience and 3.1\% by participants with high robot experience (-6.7\%). On the other hand, participants with high robot experience asked more questions, for example, in the categories potential-issues (+4.6\%), how-did-you-decide (+3.5\%), mental-state (+3.1\%), and what-abilities (+2.9\%). 

\section{Discussion and Conclusion}

In this paper, we presented a dataset of 1,893 user questions that can help robot practitioners understand user needs when designing natural language interfaces. Below, we discuss the three main findings of (i) what question categories arise, (ii) what their relative importance is, and (iii) how the question types and importance scores of different user groups vary.

\subsection{Question Categories}
We classified our data into 12 categories and 70 subcategories. This helps us understand what types of questions to expect from users. Knowing these categories also helps with understanding what type of data the robot needs to be able to access and collect during task execution. For instance, to answer questions about the environmental state, such data needs to be recorded during task execution. Beyond scanning the environment, this could even include reserving additional actions for exploration, such as checking on the location of objects.  
To answer questions about its capabilities, the robot needs access to a hard-coded capability list or a predictive model that allows it to predict what tasks it can achieve.    
To answer questions about its task performance, the robot needs to understand the users' expectations and have reference points for its usual performance. In such a manner, each category requires its own considerations of what data the robot needs to collect and have access to in order to answer the questions successfully. In our follow-up work \cite{wachowiak2026neurosymbolic}, we use the question categories to benchmark the retrieval performance of a new question-answering system, illustrating its value beyond descriptive analysis.  

Regarding the question categories, we noticed that the types of questions elicited by the text and video scenarios differ. Execution details questions were much more prevalent in the text-based scenarios (36\% of all questions) than in the videos (16\% of all questions). This is because the text descriptions provide only a high-level overview of the robot's activities, whereas video scenarios already show the details of an execution. In contrast, the videos lead to more questions on the robot's abilities and mental state. 
Text and video, as two distinct modalities through which users can be informed about robot behavior, offer another interesting and novel aspect of our dataset. The two modalities can serve as a starting point for future investigations and inform XAI-component developers with respect to their envisioned use case. For example, a use case in which people ask robots questions on-the-fly during collaborations might elicit user expectations and needs more similar to the video stimuli. In contrast, a supervising user who returns to a robot after it has completed a set of tasks might be in a situation more similar to the text case. 

\subsection{Question Importance Scores}
For each question, we also asked the participants how important it is for a robot to be able to answer that question. Through pairwise contrasts between the estimated marginal means from a linear mixed-effects model, we found various statistical differences in importance scores between question categories. No category scored below 3.5 points on the 1--5 importance score Likert scale. However, some categories were ranked clearly higher than others. For example, the most important categories dealt with potential issues the robot might run into and how it would handle them. Another important category was questions through which users tried to understand how a robot knows things, such as what to do or what an object's properties are.
In contrast, why-questions and questions about the robot's mental state (e.g., beliefs, desires) ranked lowly. When designing XAI components and natural language interfaces for robots, these findings can help with what question-answering capabilities to prioritize.

\subsection{Differences between Participant Groups}
In the last part of our exploratory data analysis, we found that participants' personal and societal-level attitudes towards robots correlated with the average importance score they gave to their questions. This correlation suggests that individuals with higher attitudes towards robots also consider it more important for a robot to be able to answer their questions. A possible explanation for this is that people with higher attitudes towards robots also have higher expectations regarding what a robot is capable of. 
When exploring whether robot novices ask different questions than participants more experienced with robots, we find that novices tend to ask more questions in the categories execution-details and environment-state, which are both categories concerned with simple facts about what happened and the current state of things. 
This finding highlights the importance of considering one's users when designing a question-answering module, as different users will ask different questions. 

\subsection{Limitations and Future Work}

\paragraph{Ambiguous User Intent.} As mentioned in Section \ref{sec:category_coding}, interpreting some questions was occasionally challenging, and assigning them to a single category was not always straightforward. Sometimes, the annotator had to infer what the participant intended with their question. 
For example, the question ``Are you not afraid the plate will fall from the kitchen counter if you place it on the edge like that?'' could be an inquiry into the robot's assessment of the situation or be intended for the robot to correct its behavior.
To remove this uncertainty, future extensions of this dataset could also ask the user to provide an example answer for their question, which they would deem satisfying.

\paragraph{Single Annotator} The category creation and annotations were conducted by a single annotator. While the annotator conducted two entire passes over the dataset to ensure consistency, created a codebook, and worked with a second annotator to confirm the reliability of the codes, the resulting categories are necessarily shaped by the main annotator's interpretation. Different annotation schemes are conceivable, for instance, with a focus on the robot's functions, and make for interesting future research. 

\paragraph{Ecological Validity.} Another limitation of our paper is the lower ecological validity of using video and text stimuli compared to putting users into real-world human--robot interactions. However, video and text stimuli also offer crucial advantages. They allow for highly controlled conditions and a reproducible setup. Moreover, they allow us to test user reactions to a very diverse set of scenarios, showcasing different types of robots, environments, and tasks.
Lastly, data collection is much more scalable than what would be possible in the lab. Therefore, real-world studies would not replace our approach but represent an exciting avenue for complementary future research.

\paragraph{Novelty Effect.} 
While participants see multiple stimuli depicting robots executing tasks, they only see the execution of a specific task by a specific robot once. However, it may be the case that users have different questions during the initial interactions (e.g., asking the robot about its abilities) compared to later, repeated interactions (e.g., asking the robot about deviations from its normal behavior).  
In future work, we would therefore like to study how users' questions change over time as they repeatedly interact with the same robot. Given that realistic home deployments involve repeated, long-term interactions, such a study is highly relevant.

\paragraph{Stimuli.} In future research, we would like to extend the already rich set of stimuli used to elicit user questions. First, this could include longer videos (e.g., 3--5 minutes). Second, the dataset would benefit from further videos involving humans who are involved collaboratively in the tasks, as right now, humans are largely relegated to simply observing the robot or receiving an object. However, the video in which the robot feeds a cat shows that users have many questions about the interaction between pet and robot --- indicating that more interactive scenarios involving humans are likely to elicit an even broader and more diverse set of questions. Third, while not the focus of this paper, we would like to include videos of interaction ruptures, such as the robot making an error or violating a social norm. 

\paragraph{Participant Diversity.} While our participant sample was diverse in gender, age, and background, participants mostly came from Western and Anglophone countries due to Prolific's participant pool and our filter for fluency in English. Future research has the opportunity to investigate cultural differences and extend the corpus to include multilingual questions. An example of national differences in XAI research is given by Kopecka \cite{kopecka2024thesis}, who found differences in the type of people who preferred goal vs. belief-based explanations between the UK and South Korea.

\paragraph{Task-Specific Analysis} Given the trend towards general-purpose robots, the current paper analyzes user questions for household robots in a task-agnostic way. Nevertheless, follow-up studies could identify whether questions differ based on task type. Such analysis would be facilitated by each question in the dataset already being annotated with the stimulus that evoked it. Such an analysis would be especially helpful for roboticists with highly specific use cases in mind and for whom the otherwise large amount of subcategories might be difficult to apply.


\begin{acks}
Lennart Wachowiak was supported by the Center for Doctoral Training (CDT) in Safe and Trusted AI (STAI) and UKRI (EP/S023356/1), as well as the King’s Institute for Artificial Intelligence. We thank Michelle Nwachukwu for helping with the inter-annotator agreement experiment.  
\end{acks}

\bibliographystyle{ACM-Reference-Format}
\bibliography{sample-base}

@inproceedings{mcguinness2007categorization,
  title={{A Categorization of Explanation Questions for Task Processing Systems}},
  author={McGuinness, Deborah L and Glass, Alyssa and Wolverton, Michael and Da Silva, Paulo Pinheiro},
  booktitle={ExaCt},
  pages={42--48},
  year={2007}
}

@inproceedings{wachowiak2024people,
  title={{When do People Want an Explanation from a Robot?}},
  author={Wachowiak, Lennart and Fenn, Andrew and Kamran, Haris and Coles, Andrew and Celiktutan, Oya and Canal, Gerard},
  booktitle={ACM/IEEE International Conference on Human-Robot Interaction (HRI)},
  pages={752--761},
  year={2024}
}

@inproceedings{padmakumar2022teach,
  title={{TEACh: Task-Driven Embodied Agents that Chat}},
  author={Padmakumar, Aishwarya and Thomason, Jesse and Shrivastava, Ayush and Lange, Patrick and Narayan-Chen, Anjali and Gella, Spandana and Piramuthu, Robinson and Tur, Gokhan and Hakkani-Tur, Dilek},
  booktitle={Proceedings of the AAAI Conference on Artificial Intelligence},
  volume={36},
  number={2},
  pages={2017--2025},
  year={2022}
}

@article{gao2022dialfred,
  title={{DialFRED: Dialogue-Enabled Agents for Embodied Instruction Following}},
  author={Gao, Xiaofeng and Gao, Qiaozi and Gong, Ran and Lin, Kaixiang and Thattai, Govind and Sukhatme, Gaurav S},
  journal={IEEE Robotics and Automation Letters},
  volume={7},
  number={4},
  pages={10049--10056},
  year={2022},
  publisher={IEEE}
}

@inproceedings{shridhar2020alfred,
  title={{ALFRED: A Benchmark for Interpreting Grounded Instructions for Everyday Tasks}},
  author={Shridhar, Mohit and Thomason, Jesse and Gordon, Daniel and Bisk, Yonatan and Han, Winson and Mottaghi, Roozbeh and Zettlemoyer, Luke and Fox, Dieter},
  booktitle={Proceedings of the IEEE/CVF conference on computer vision and pattern recognition},
  pages={10740--10749},
  year={2020}
}

@inproceedings{bastianelli2014huric,
  title={{HuRIC: a Human Robot Interaction Corpus}},
  author={Bastianelli, Emanuele and Castellucci, Giuseppe and Croce, Danilo and Iocchi, Luca and Basili, Roberto and Nardi, Daniele and others},
  booktitle={LREC},
  pages={4519--4526},
  year={2014}
}

@inproceedings{nair2022learning,
  title={{Learning Language-Conditioned Robot Behavior from Offline Data and Crowd-Sourced Annotation}},
  author={Nair, Suraj and Mitchell, Eric and Chen, Kevin and Savarese, Silvio and Finn, Chelsea and others},
  booktitle={Conference on Robot Learning},
  pages={1303--1315},
  year={2022},
  organization={PMLR}
}

@article{ji2019survey,
  title={{A Survey of Human Action Analysis in HRI Applications}},
  author={Ji, Yanli and Yang, Yang and Shen, Fumin and Shen, Heng Tao and Li, Xuelong},
  journal={IEEE Transactions on Circuits and Systems for Video Technology},
  volume={30},
  number={7},
  pages={2114--2128},
  year={2019},
  publisher={IEEE}
}

@inproceedings{gucsi2025hri,
  title={{HRI-SENSE: A Multimodal Dataset on Social and Emotional Responses to Robot Behaviour}},
  author={Gucsi, Balint and Tuyen, Nguyen Tan Viet and Chu, Bing and Tarapore, Danesh and Tran-Thanh, Long},
  booktitle={2025 20th ACM/IEEE International Conference on Human-Robot Interaction (HRI)},
  pages={1319--1323},
  year={2025},
  organization={IEEE}
}

@inproceedings{candon2024react,
  title={{REACT: Two Datasets for Analyzing Both Human Reactions and Evaluative Feedback to Robots over Time}},
  author={Candon, Kate and Georgiou, Nicholas C and Zhou, Helen and Richardson, Sidney and Zhang, Qiping and Scassellati, Brian and V{\'a}zquez, Marynel},
  booktitle={Proceedings of the 2024 ACM/IEEE International Conference on Human-Robot Interaction},
  pages={885--889},
  year={2024}
}

@inproceedings{spitale2024err,
  title={{Err@HRI 2024 Challenge: Multimodal Detection of Errors and Failures in Human-Robot Interactions}},
  author={Spitale, Micol and Parreira, Maria Teresa and Stiber, Maia and Axelsson, Minja and Kara, Neval and Kankariya, Garima and Huang, Chien-Ming and Jung, Malte and Ju, Wendy and Gunes, Hatice},
  booktitle={Proceedings of the 26th International Conference on Multimodal Interaction},
  pages={652--656},
  year={2024}
}

@article{ko2021air,
  title={{AIR-Act2Act: Human--human interaction dataset for teaching non-verbal social behaviors to robots}},
  author={Ko, Woo-Ri and Jang, Minsu and Lee, Jaeyeon and Kim, Jaehong},
  journal={The International Journal of Robotics Research},
  volume={40},
  number={4-5},
  pages={691--697},
  year={2021},
  publisher={SAGE Publications Sage UK: London, England}
}

@inproceedings{li2023behavior,
  title={{Behavior-1K: A Benchmark for Embodied AI with 1,000 Everyday Activities and Realistic Simulation}},
  author={Li, Chengshu and Zhang, Ruohan and Wong, Josiah and Gokmen, Cem and Srivastava, Sanjana and Mart{\'\i}n-Mart{\'\i}n, Roberto and Wang, Chen and Levine, Gabrael and Lingelbach, Michael and Sun, Jiankai and others},
  booktitle={Conference on Robot Learning},
  pages={80--93},
  year={2023},
  organization={PMLR}
}

@inproceedings{wachowiak2026neurosymbolic,
  title={{Neurosymbolic Explanation Selection in Robotics: Combining the Strengths of Planning and Foundation Models for XAI}},
  author={Wachowiak, Lennart and Coles, Andrew I and Celiktutan, Oya and Canal, Gerard},
  booktitle={Companion Proceedings of the 21st ACM/IEEE International Conference on Human-Robot Interaction},
  pages={222--227},
  year={2026}
}

@article{team2025gemini,
  title={{Gemini Robotics: Bringing AI into the Physical World}},
  author={Team, Gemini Robotics and Abeyruwan, Saminda and Ainslie, Joshua and Alayrac, Jean-Baptiste and Arenas, Montserrat Gonzalez and Armstrong, Travis and Balakrishna, Ashwin and Baruch, Robert and Bauza, Maria and Blokzijl, Michiel and others},
  journal={arXiv preprint arXiv:2503.20020},
  year={2025}
}

@inproceedings{fu2024mobile,
  title={{Mobile ALOHA: Learning Bimanual Mobile Manipulation Using Low-Cost Whole-Body Teleoperation}},
  author={Fu, Zipeng and Zhao, Tony Z and Finn, Chelsea},
year = {2024},
  booktitle={8th Annual Conference on Robot Learning (CoRL)}
}

@article{intelligence2025pi_,
  title={{A Vision-Language-Action Model with Open-World Generalization}},
  author={Intelligence, Physical and Black, Kevin and Brown, Noah and Darpinian, James and Dhabalia, Karan and Driess, Danny and Esmail, Adnan and Equi, Michael and Finn, Chelsea and Fusai, Niccolo and others},
  journal={arXiv preprint arXiv:2504.16054},
  year={2025}
}

@article{koverola2022general,
  title={{General Attitudes Towards Robots Scale (GAToRS): A New Instrument for Social Surveys}},
  author={Koverola, Mika and Kunnari, Anton and Sundvall, Jukka and Laakasuo, Michael},
  journal={International Journal of Social Robotics},
  volume={14},
  number={7},
  pages={1559--1581},
  year={2022},
  publisher={Springer}
}

@article{wachowiak2024taxonomy,
  title={{A Taxonomy of Explanation Types and Need Indicators in Human--Agent Collaborations}},
  author={Wachowiak, Lennart and Coles, Andrew and Canal, Gerard and Celiktutan, Oya},
  journal={International Journal of Social Robotics},
  volume={16},
  number={7},
  pages={1681--1692},
  year={2024},
  publisher={Springer}
}

@article{raza2022wild,
  title={{An In-the-Wild Study to Find Type of Questions People Ask to a Social Robot Providing Question-Answering Service}},
  author={Raza, Syed Ali and Vitale, Jonathan and Tonkin, Meg and Johnston, Benjamin and Billingsley, Richard and Herse, Sarita and Williams, Mary-Anne},
  journal={Intelligent Service Robotics},
  volume={15},
  number={3},
  pages={411--426},
  year={2022},
  publisher={Springer}
}

@article{huang2025survey,
  title={{A Survey on Hallucination in Large Language Models: Principles, Taxonomy, Challenges, and Open Questions}},
  author={Huang, Lei and Yu, Weijiang and Ma, Weitao and Zhong, Weihong and Feng, Zhangyin and Wang, Haotian and Chen, Qianglong and Peng, Weihua and Feng, Xiaocheng and Qin, Bing and others},
  journal={Transactions on Information Systems},
  volume={43},
  number={2},
  pages={1--55},
  year={2025},
  publisher={ACM}
}

@inproceedings{kopecka2020explainable,
  title={{Explainable AI for Cultural Minds}},
  author={Kopecka, Hana and Such, Jose},
  booktitle={Workshop on Dialogue, Explanation and Argumentation for Human--Agent Interaction},
  year={2020}
}

@article{miller2019explanation,
  title={{Explanation in Artificial Intelligence: Insights from the Social Sciences}},
  author={Miller, Tim},
  journal={Artificial intelligence},
  volume={267},
  year={2019},
  publisher={Elsevier}
}

@inproceedings{lemasurier2024templated,
  title={Templated vs. Generative: Explaining Robot Failures},
  author={LeMasurier, Gregory and Tagliamonte, Christian and Breen, Jacob and Maccaline, Daniel and Yanco, Holly A},
  booktitle={2024 33rd IEEE International Conference on Robot and Human Interactive Communication (ROMAN)},
  pages={1346--1353},
  year={2024},
  publisher={IEEE}
}

@inproceedings{tagliamonte2024generalizable,
  title={{A Generalizable Architecture for Explaining Robot Failures Using Behavior Trees and Large Language Models}},
  author={Tagliamonte, Christian and Maccaline, Daniel and LeMasurier, Gregory and Yanco, Holly A},
  booktitle={Companion of the ACM/IEEE International Conference on Human-Robot Interaction (HRI)},
  pages={1038--1042},
  year={2024}
}

@book{marr2010vision,
  title={Vision: A computational investigation into the human representation and processing of visual information},
  author={Marr, David},
  year={1982},
}

@inproceedings{fox2017explainable,
	title        = {{Explainable Planning}},
	author       = {Fox, Maria and Long, Derek and Magazzeni, Daniele},
	year         = {2017},
	booktitle    = {Workshop on Explainable Planning at the International Joint Conference on Artificial Intelligence (IJCAI)}
}

@article{buhrmester2021analysis,
	title        = {{Analysis of Explainers of Black Box Deep Neural Networks for Computer Vision: A Survey}},
	author       = {Buhrmester, Vanessa and M{\"u}nch, David and Arens, Michael},
	year         = {2021},
	journal      = {Machine Learning and Knowledge Extraction},
	volume       = {3},
pages = {966--989}
}

@ARTICLE{8466590,
  author={Adadi, Amina and Berrada, Mohammed},
  journal={IEEE Access}, 
  title={{Peeking Inside the Black-Box: A Survey on Explainable Artificial Intelligence (XAI)}}, 
  year={2018},
  volume={6},
  number={},
  pages={52138-52160}}

@article{setchi2020explainable,
	title        = {Explainable robotics in human-robot interactions},
	author       = {Setchi, Rossitza and Dehkordi, Maryam Banitalebi and Khan, Juwairiya Siraj},
	year         = {2020},
	journal      = {Procedia Computer Science},
	volume       = {176},
	pages        = {3057--3066}
}

@phdthesis{kopecka2024thesis,
  title={{Preferences for AI Explanations: Considering the Role of User and Robot
Characteristics}},
  author={Kopecka, Hana},
school={King's College London},
  year={2024}
}

@inproceedings{jayagopi2013vernissage,
  title={{The Vernissage Corpus: A Conversational Human-Robot-Interaction Dataset}},
  author={Jayagopi, Dinesh Babu and Sheiki, Samira and Klotz, David and Wienke, Johannes and Odobez, Jean-Marc and Wrede, Sebastien and Khalidov, Vasil and Nyugen, Laurent and Wrede, Britta and Gatica-Perez, Daniel},
  booktitle={ACM/IEEE International Conference on Human-Robot Interaction (HRI)},
  pages={149--150},
  year={2013},
  publisher={IEEE}
}

@inproceedings{lukin-etal-2024-scout,
    title = "{SCOUT}: A Situated and Multi-Modal Human-Robot Dialogue Corpus",
    author = "Lukin, Stephanie M.  and
      Bonial, Claire  and
      Marge, Matthew  and
      Hudson, Taylor A.  and
      Hayes, Cory J.  and
      Pollard, Kimberly  and
      Baker, Anthony  and
      Foots, Ashley N.  and
      Artstein, Ron  and
      Gervits, Felix  and
      Abrams, Mitchell  and
      Henry, Cassidy  and
      Donatelli, Lucia  and
      Leuski, Anton  and
      Hill, Susan G.  and
      Traum, David  and
      Voss, Clare",
    editor = "Calzolari, Nicoletta  and
      Kan, Min-Yen  and
      Hoste, Veronique  and
      Lenci, Alessandro  and
      Sakti, Sakriani  and
      Xue, Nianwen",
    booktitle = "Proceedings of the Joint International Conference on Computational Linguistics, Language Resources and Evaluation (LREC-COLING)",
    year = "2024",
    publisher = "ELRA and ICCL",
    pages = "14445--14458",
}

@inproceedings{wachowiak2023survey,
  title={{A Survey of Evaluation Methods and Metrics for Explanations in Human--Robot Interaction (HRI)}},
  author={Wachowiak, Lennart and Celiktutan, Oya and Coles, Andrew and Canal, Gerard},
  booktitle={Explainable Robotics Workshop at IEEE International Conference on Robotics and Automation (ICRA)},
  year={2023}
}

@inproceedings{eiband2018bringing,
  title={Bringing transparency design into practice},
  author={Eiband, Malin and Schneider, Hanna and Bilandzic, Mark and Fazekas-Con, Julian and Haug, Mareike and Hussmann, Heinrich},
  booktitle={Conference on Intelligent User Interfaces},
publisher = {ACM},
  year={2018}
}

@inproceedings{ehsan10.1007/978-3-030-60117-1_33,
author = {Ehsan, Upol and Riedl, Mark O.},
title = {{Human-Centered Explainable AI: Towards a Reflective Sociotechnical Approach}},
year = {2020},
isbn = {978-3-030-60116-4},
publisher = {Springer},
booktitle = {HCI International - Late Breaking Papers},
}

@inproceedings{mucha2020towards,
  title={{Towards Participatory Design Spaces for Explainable AI Interfaces in Expert Domains}},
  author={Mucha, Henrik and Robert, Sebastian and Breitschwerdt, R{\"u}diger and Fellmann, Michael},
  booktitle={German Conference on AI},
  year={2020}
}

@inproceedings{gebelli2024co,
  title={{Co-Designing Explainable Robots: A Participatory Design Approach for HRI}},
  author={Gebell{\'\i}, Ferran and Ros, Raquel and Lemaignan, S{\'e}verin and Garrell, Ana{\'\i}s},
  booktitle={IEEE International Conference on Robot and Human Interactive Communication (ROMAN)},
  pages={1564--1570},
  year={2024},
  publisher={IEEE}
}

@INPROCEEDINGS{wangtrust2016,

  author={Wang, Ning and Pynadath, David V. and Hill, Susan G.},

  booktitle={International Conference on Human-Robot Interaction (HRI)}, 

  title={{Trust Calibration within a Human--Robot Team: Comparing Automatically Generated Explanations}}, 

  year={2016},

  volume={},

  number={},

  pages={109--116},
}

@article{nielsen2023using,
  title={{Using User-Generated YouTube Videos to Understand Unguided Interactions with Robots in Public Places}},
  author={Nielsen, Sara and Skov, Mikael B and Hansen, Karl Damkj{\ae}r and Kaszowska, Aleksandra},
  journal={ACM Transactions on Human-Robot Interaction},
  volume={12},
  number={1},
  pages={1--40},
  year={2023},
  publisher={ACM}
}

@misc{aisi2024-should-ai-behave,
  author       = {{UK AI Security Institute}},
  title        = {{Should AI Systems Behave Like People?}},
  year         = {2024},
  month        = sep,
  day          = {25},
  url          = {https://www.aisi.gov.uk/work/should-ai-systems-behave-like-people},
  note         = {Accessed 2025-09-16}
}

@inproceedings{9562003,
	title        = {{Towards Providing Explanations for Robot Motion Planning}},
	author       = {Brandão, Martim and Canal, Gerard and Krivi{\'c}, Senka and Magazzeni, Daniele},
	year         = {2021},
	booktitle    = {International Conference on Robotics and Automation (ICRA)},
	pages        = {3927--3933},
	organization = {IEEE}
}

@article{comanici2025gemini,
  title={{Gemini 2.5: Pushing the Frontier with Advanced Reasoning, Multimodality, Long Context, and Next Generation Agentic Capabilities}},
  author={Comanici, Gheorghe and Bieber, Eric and Schaekermann, Mike and Pasupat, Ice and Sachdeva, Noveen and Dhillon, Inderjit and Blistein, Marcel and Ram, Ori and Zhang, Dan and Rosen, Evan and others},
  journal={arXiv preprint arXiv:2507.06261},
  year={2025}
}

@article{trainum2023robots,
  title={{Robots in Assisted Living Facilities: Scoping Review}},
  author={Trainum, Katie and Tunis, Rachel and Xie, Bo and Hauser, Elliott},
  journal={JMIR Aging},
  volume={6},
  number={1},
  pages={e42652},
  year={2023},
  publisher={JMIR Publications}
}

@article{gonzalez2021service,
  title={{Service Robots: Trends and Technology}},
  author={Gonzalez-Aguirre, Juan Angel and Osorio-Oliveros, Ricardo and Rodr{\'\i}guez-Hern{\'a}ndez, Karen L and Liz{\'a}rraga-Iturralde, Javier and Morales Menendez, Ruben and Ramirez-Mendoza, Ricardo A and Ramirez-Moreno, Mauricio Adolfo and Lozoya-Santos, Jorge de Jesus},
  journal={Applied Sciences},
  volume={11},
  number={22},
  pages={10702},
  year={2021},
  publisher={MDPI}
}

@inproceedings{anjomshoae2019explainable,
	title        = {{Explainable Agents and Robots: Results from a Systematic Literature Review}},
	author       = {Anjomshoae, Sule and Najjar, Amro and Calvaresi, Davide and Fr\"{a}mling, Kary},
	year         = {2019},
	booktitle    = {International Conference on Autonomous Agents and Multiagent Systems (AAMAS)},
	pages        = {1078–1088},
	isbn         = {9781450363099},
	numpages     = {11},
	organization = {IFAAMAS}
}

@inproceedings{wang2016impact,
	title        = {{The Impact of POMDP-Generated Explanations on Trust and Performance in Human-Robot Teams}},
	author       = {Wang, Ning and Pynadath, David V. and Hill, Susan G.},
	year         = {2016},
	booktitle    = {International Conference on Autonomous Agents and Multiagent Systems (AAMAS)},
	publisher    = {IFAAMAS},
	pages        = {997–1005},
	isbn         = {9781450342391},
	numpages     = {9},
}

@article{lyons2023explanations,
  title={{Explanations and Trust: What Happens to Trust When a Robot Partner Does Something Unexpected?}},
  author={Lyons, Joseph B and aldin Hamdan, Izz and Vo, Thy Q},
  journal={Computers in Human Behavior},
  volume={138},
  pages={107473},
  year={2023},
  publisher={Elsevier}
}

@article{norman2010likert,
  title={{Likert Scales, Levels of Measurement and the “Laws” of Statistics}},
  author={Norman, Geoff},
  journal={Advances in health sciences education},
  volume={15},
  number={5},
  pages={625--632},
  year={2010},
  publisher={Springer}
}

@article{choi2021err,
	title        = {{To Err Is Human(-oid): How Do Consumers React to Robot Service Failure and Recovery?}},
	author       = {Sungwoo Choi and Anna S. Mattila and Lisa E. Bolton},
	year         = {2021},
	journal      = {Journal of Service Research},
	volume       = {24},
	number       = {3},
	pages        = {354--371}
}

@article{fischer2018increasing,
  title={{Increasing trust in human--robot medical interactions: effects of transparency and adaptability}},
  author={Fischer, Kerstin and Weigelin, Hanna Mareike and Bodenhagen, Leon},
  journal={Paladyn, Journal of Behavioral Robotics},
  volume={9},
  number={1},
  pages={95--109},
  year={2018},
  publisher={Sciendo}
}

@article{doshi2017towards,
  title={{Towards a Rigorous Science of Interpretable Machine Learning}},
  author={Doshi-Velez, Finale and Kim, Been},
  journal={arXiv preprint arXiv:1702.08608},
  year={2017}
}

@book{montgomery1982introduction,
  title={Introduction to linear regression analysis},
  author={Montgomery, DC and Peck, EA},
  year={1982},
  publisher={John Wiley \& Sons},
address={New York}
}

@article{lindstrom1988newton,
  title={{Newton—Raphson and EM algorithms for Linear Mixed-Effects Models for Repeated-Measures Data}},
  author={Lindstrom, Mary J and Bates, Douglas M},
  journal={Journal of the American Statistical Association},
  volume={83},
  number={404},
  pages={1014--1022},
  year={1988},
  publisher={Taylor \& Francis}
}

@inproceedings{gebelli2025personalised,
  title={Personalised Explainable Robots Using LLMs},
  author={Gebell{\'\i}, Ferran and Hriscu, Lavinia and Ros, Raquel and Lemaignan, S{\'e}verin and Sanfeliu, Alberto and Garrell, Ana{\'\i}s},
  booktitle={ACM/IEEE International Conference on Human-Robot Interaction (HRI)},
  pages={1304--1308},
  year={2025},
  publisher={IEEE}
}

@inproceedings{wachowiak2024large,
  title={{Are Large Language Models Aligned with People’s Social Intuitions for Human--Robot Interactions?}},
  author={Wachowiak, Lennart and Coles, Andrew and Celiktutan, Oya and Canal, Gerard},
  booktitle={IEEE/RSJ International Conference on Intelligent Robots and Systems (IROS)},
  pages={2520--2527},
  year={2024},
  publisher={IEEE}
}

@article{sobrin2024explaining,
  title={Explaining Autonomy: Enhancing Human-Robot Interaction through Explanation Generation with Large Language Models},
  author={Sobr{\'\i}n-Hidalgo, David and Gonz{\'a}lez-Santamarta, Miguel A and Guerrero-Higueras, {\'A}ngel M and Rodr{\'\i}guez-Lera, Francisco J and Matell{\'a}n-Olivera, Vicente},
  journal={arXiv preprint arXiv:2402.04206},
  year={2024}
}

@inproceedings{cashmore2019towards,
	title        = {{Towards Explainable AI Planning as a Service}},
	author       = {Cashmore, Michael and Collins, Anna and Krarup, Benjamin and Krivic, Senka and Magazzeni, Daniele and Smith, David},
	year         = {2019},
	booktitle      = {ICAPS Workshop XAIP}
}

@article{landisMeasurementObserverAgreement1977,
  title = {The {{Measurement}} of {{Observer Agreement}} for {{Categorical Data}}},
  author = {Landis, J. Richard and Koch, Gary G.},
  year = {1977},
  journal = {Biometrics},
  volume = {33},
  number = {1},
  issn = {0006341X},
  langid = {english}
}

@ARTICLE{2020SciPy,
  author  = {Virtanen, Pauli and Gommers, Ralf and Oliphant, Travis E. and
            Haberland, Matt and Reddy, Tyler and Cournapeau, David and
            Burovski, Evgeni and Peterson, Pearu and Weckesser, Warren and
            Bright, Jonathan and {van der Walt}, St{\'e}fan J. and
            Brett, Matthew and Wilson, Joshua and Millman, K. Jarrod and
            Mayorov, Nikolay and Nelson, Andrew R. J. and Jones, Eric and
            Kern, Robert and Larson, Eric and Carey, C J and
            Polat, {\.I}lhan and Feng, Yu and Moore, Eric W. and
            {VanderPlas}, Jake and Laxalde, Denis and Perktold, Josef and
            Cimrman, Robert and Henriksen, Ian and Quintero, E. A. and
            Harris, Charles R. and Archibald, Anne M. and
            Ribeiro, Ant{\^o}nio H. and Pedregosa, Fabian and
            {van Mulbregt}, Paul and {SciPy 1.0 Contributors}},
  title   = {{{SciPy} 1.0: Fundamental Algorithms for Scientific
            Computing in Python}},
  journal = {Nature Methods},
  year    = {2020},
  volume  = {17},
  pages   = {261--272},
}

@article{holm1979simple,
 ISSN = {03036898, 14679469},
 author = {Sture Holm},
 journal = {Scandinavian Journal of Statistics},
 number = {2},
 pages = {65--70},
 publisher = {[Board of the Foundation of the Scandinavian Journal of Statistics, Wiley]},
 title = {{A Simple Sequentially Rejective Multiple Test Procedure}},
 urldate = {2023-11-22},
 volume = {6},
 year = {1979}
}

@article{kopecka2024preferences,
  title={{Preferences for AI Explanations Based on Cognitive Style and Socio-Cultural Factors}},
  author={Kopecka, Hana and Such, Jose and Luck, Michael},
  journal={Proceedings of the ACM on Human-Computer Interaction},
  volume={8},
  number={CSCW1},
  pages={1--32},
  year={2024},
  publisher={ACM}
}

@article{spearman1904proof,
  title={{The Proof and Measurement of Association between Two Things}},
  author={Spearman, C},
  journal={The American Journal of Psychology},
  volume={15},
  number={1},
  pages={72--101},
  year={1904}
}

@inproceedings{seabold2010statsmodels,
  title={{statsmodels: Econometric and Statistical Modeling with Python}},
  author={Seabold, Skipper and Perktold, Josef},
  booktitle={Python in Science Conference},
  year={2010},
}

@inproceedings{esterwood2022having,
  title={{Having the Right Attitude: How Attitude Impacts Trust Repair in Human—Robot Interaction}},
  author={Esterwood, Connor and Robert, Lionel P},
  booktitle={ACM/IEEE International Conference on Human-Robot Interaction (HRI)},
  pages={332--341},
  year={2022},
  publisher={IEEE}
}

@article{guo2025deepseek,
  title={{Deepseek-R1: Incentivizing Reasoning Capability in LLMs via Reinforcement Learning}},
  author={Guo, Daya and Yang, Dejian and Zhang, Haowei and Song, Junxiao and Zhang, Ruoyu and Xu, Runxin and Zhu, Qihao and Ma, Shirong and Wang, Peiyi and Bi, Xiao and others},
  journal={arXiv preprint arXiv:2501.12948},
  year={2025}
}

@article{achiam2023gpt,
  title={{GPT-4 Technical Report}},
  author={Achiam, Josh and Adler, Steven and Agarwal, Sandhini and Ahmad, Lama and Akkaya, Ilge and Aleman, Florencia Leoni and Almeida, Diogo and Altenschmidt, Janko and Altman, Sam and Anadkat, Shyamal and others},
  journal={arXiv preprint arXiv:2303.08774},
  year={2023}
}

\appendix

\section{Online Resources} \label{app:online}

We made the data and code available on GitHub: \url{https://github.com/lwachowiak/xai-questions-dataset}.
The repository contains:
\begin{itemize}
    \item the dataset of all natural language questions, their categories, and their importance scores
    \item our analysis code
    \item the annotations from a second annotator for computing the inter-annotator agreement
    \item the questionnaire form to ensure the reproducibility of our work
\end{itemize}

Additionally, the dataset is hosted on \href{https://huggingface.co/datasets/lwachowiak/xai-questions-dataset}{Hugging Face}.

\section{Video Stimuli References} \label{app:video-ref}

The frames presented in Table \ref{tab:video_overview} and the corresponding video stimuli are snippets from the YouTube videos linked below. If a video contains multiple tasks, we indicate the task used for our video stimulus by providing the relevant timestamp. For our stimuli, we sometimes changed the playback speed and may have removed elements such as teleoperators and text boxes through cropping. To see the exact stimuli used, contact the corresponding author of this paper.
\begin{itemize}
     \item {[V1] \url{https://www.youtube.com/watch?v=1r2GqaGyyIA}, Gemini Robotics --- Salad Making}
    \item {[V2] \url{https://www.youtube.com/watch?v=ikZeU3wKVjM}, Gemini Robotics --- Lunchbox (task starting at 0:50)}
    \item {[V3] \url{https://www.youtube.com/watch?v=Z3yQHYNXPws}, Figure --- Fridge Sorting}
    \item  {[V4] \url{https://www.youtube.com/watch?v=Zn8yMaepzVk}, Physical Intelligence --- Kitchen (task starting at 0:34)} 
    \item  {[V5] \url{https://www.youtube.com/watch?v=Zn8yMaepzVk}, Physical Intelligence --- Bedroom (task starting at 1:23)} 
    \item {[V6] \url{https://www.youtube.com/watch?v=L3rLT84qqLk}, Astribot --- Drawer Sorting (task starting at 1:23)}
    \item {[V7] \url{https://www.youtube.com/watch?v=2Fx5hBvT0Ac}, Physical Intelligence --- Dryer Unpacking}
    \item {[V8] \url{https://www.youtube.com/watch?v=Oa19cq_MxE0}, Physical Intelligence --- Laundry Folding}
    \item {[V9] \url{https://www.youtube.com/watch?v=XpBWxLg-3bI}, 1X --- Shirt-Handover (task starting at 0:38)}
    \item {[V10] \url{https://www.youtube.com/watch?v=Ni4p8axgqHM}, Hello Robot --- Dishwasher (task starting at 0:34)}
    \item  {[V11] \url{https://www.youtube.com/watch?v=6X-s4Qsn1z4}, Astribot --- Cat Feeding (task starting at 0:18)}
    \item  {[V12] \url{https://www.youtube.com/watch?v=6X-s4Qsn1z4}, Astribot --- Waffle Making (task starting at 0:47)}
    \item  {[V13] \url{https://www.youtube.com/watch?v=6X-s4Qsn1z4}, Astribot --- Tea Preparation (task starting at 1:53)}
    \item  {[V14] \url{https://www.youtube.com/watch?v=mnLVbwxSdNM}, Mobile ALOHA --- Omelette (task starting at 0:11)}
    \item  {[V15] \url{https://www.youtube.com/watch?v=mnLVbwxSdNM}, Mobile ALOHA --- Cooking (task starting at 0:34)}
\end{itemize}
Our use of the videos falls under YouTube's fair use policy\footnote{Fair use on YouTube (Retrieved September 10th, 2025): \url{https://support.google.com/youtube/answer/9783148?hl=en}} that specifically mentions the use in research without the copyright owner’s permission. The frames in Table \ref{tab:video_overview} do not show any identifiable people. Similar use of YouTube videos can be found, for example, in \cite{nielsen2023using}. 










\end{document}